\documentclass[sn-mathphys-num]{sn-jnl}

\usepackage{graphicx}%
\usepackage{multirow}%
\usepackage{amsmath,amssymb,amsfonts}%
\usepackage{amsthm}%
\usepackage{mathrsfs}%
\usepackage[title]{appendix}%
\usepackage[table,xcdraw]{xcolor}
\usepackage{textcomp}%
\usepackage{manyfoot}%
\usepackage{booktabs}%
\usepackage{algorithm}%
\usepackage{algorithmicx}%
\usepackage{algpseudocode}%
\usepackage{listings}%
\usepackage[table,xcdraw]{}%
\usepackage{array}%
\usepackage{adjustbox}%
\usepackage{lipsum}%
\usepackage{comment}%

\theoremstyle{thmstyleone}%
\newtheorem{theorem}{Theorem}
\newtheorem{proposition}[theorem]{Proposition}%

\theoremstyle{thmstyletwo}%
\newtheorem{example}{Example}%
\newtheorem{remark}{Remark}%

\theoremstyle{thmstylethree}%
\newtheorem{definition}{Definition}%

\begin{document}

\title[Article Title]{Review of Autonomous Mobile Robots for the Warehouse Environment}


\author*[1]{\fnm{Russell} \sur{Keith}}\email{rkeith@nevada.unr.edu}

\author[1]{\fnm{Hung} \sur{Manh La}}\email{hla@unr.edu}
\equalcont{These authors contributed equally to this work.}


\affil[1]{\orgdiv{Advanced Robotics and Automation (ARA) Lab}, \orgname{University of Nevada, Reno}, \orgaddress{\street{1664 N Virginia St.}, \city{Reno}, \postcode{89557}, \state{NV}, \country{U.S}}}




\abstract{Autonomous mobile robots (AMRs) have been a rapidly expanding research topic for the past decade. Unlike their counterpart, the automated guided vehicle (AGV), AMRs can make decisions and do not need any previously installed infrastructure to navigate. Recent technological developments in hardware and software have made them more feasible, especially in warehouse environments. Traditionally, most wasted warehouse expenses come from the logistics of moving material from one point to another, and is exhaustive for humans to continuously walk those distances while carrying a load. Here, AMRs can help by working with humans to cut down the time and effort of these repetitive tasks, improving performance and reducing the fatigue of their human collaborators. This literature review covers the recent developments in AMR technology including hardware, robotic control, and system control. This paper also discusses examples of current AMR producers, their robots, and the software that is used to control them. We conclude with future research topics and where we see AMRs developing in the warehouse environment.}




\maketitle

\section{Introduction}\label{sec1}
Autonomous mobile robots (AMRs) have been an active research field since the 1950s when the first Automated Guided Vehicle (AGV) was introduced~\cite{muller_automated_1983}. Recently, with the hardware advances in sensors and computing power, they have become more feasible. They have now been introduced into the manufacturing environment, particularly with logistics and material handling. In a traditional warehouse environment, goods are received on a truck and placed on a shelf. Then a team of pickers receives a list of orders that need to be filled for that day. An individual will then be assigned an order and then proceed to walk about the warehouse picking each item off a shelf and placing it in a container. Once the picker has finished the container will then be packaged, shipped, or knitted depending on the nature of the warehouse. Some companies modify this process and assign floor zones to individuals. Each person is responsible for picking items in their zone and bringing them to a sorting area.  

\begin{figure}[t]
    \centering
    \includegraphics[width=2.5in]{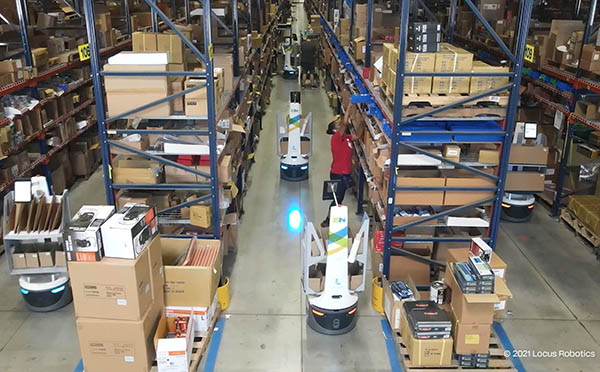}
    \caption{Human-robot collaboration in the warehouse~\cite{chain_how_nodate}}
    \label{Fig: 1a}
\end{figure}

\begin{figure}[h]
    \centering
    \includegraphics[width=3.5 in]{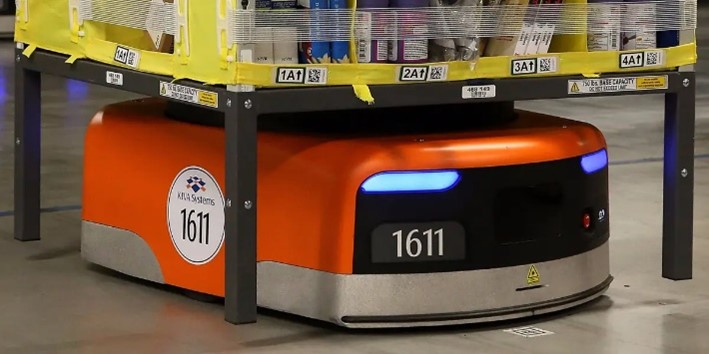}
    \caption{Amazon's Kiva robot~\cite{kim_amazon_nodate}}
    \label{Fig: 1b}
\end{figure}

Traditional warehouse structures suffer from motion waste, which is one of the 8 wastes in Lean manufacturing~\cite{longhan_production_2013}. For companies to stay competitive, and lower logistic costs, many of them have turned to “smart warehouses”, which deploy a series of tools to lower waste and increase production. Automated robots are used that either pick items or work with pickers to reduce the walking distance needed [see Fig.~\ref{Fig: 1a}]. A good example of this is Amazon with their Kiva robots [see  Fig.~\ref{Fig: 1b}] ~\cite{banker_new_nodate}. Smart warehouses are not limited to robots, they also implement better planning strategies through warehouse management software (WMS), resource management, and path planning. Converting to smart warehousing does include some downsides. For example, the initial investment is high \$150 - \$200 per square foot but, the returns from reducing waste have shown to be substantial in the longer term~\cite{kamali_smart_2019}. 

In the manufacturing environment, robot-human interactions are becoming more important in daily operations. To help with efficiency, humans work alongside robots to pick up the shortcomings of each side. In warehouses, products of different sizes or shapes are handled from shelves and placed into bins. Robots have a particularly difficult time with this task since they are limited to the end effector installed as well as figuring out how to hold the object without it falling over. Human operators can easily do this task, but they are limited to fatigue after traveling back and forth across the warehouse. A solution would be to have the robot meet the picker at the product location and have the picker move the product off the shelf and onto the AMR. This paper specifically looks at this scenario and filters the search results based on their relevance.

\begin{table}[!t]
    \centering
\caption{Reference Summary}
\label{tab1}
    \begin{tabular}{ccc} 
        \toprule
         &Subcategory&\\ 
         \midrule
         Hardware& Sensors & [7] - [21] \\ 
                 & Processors & [22] - [26] \\ 
                 & Batteries & [27] - [31] \\ 
        \toprule
         Robotic Control &  Localization & [32] - [35] \\ 
                & Artificial Intelligence & [36] - [39] \\ 
                & Path Planning & [40] - [57] \\ 
        \toprule
        System Control & Resource Management & [58] - [68] \\ 
                & Scheduling &[69] - [79] \\ 
                & Human-Robot Picking Methods & [80] - [84] \\ 
                & Warehouse Flow and Layout Design & [85] - [96] \\ 
        \toprule
        Future Work & &[97] - [107] \\ 
    \end{tabular}
\end{table}

\subsection{Review Methodology}
This paper looks at the current state of AMRs that are being used in the warehouse environment. It includes not only research papers but also covers current AMR company hardware and the adjacent software. This paper used previous literature reviews, and research articles from IEEE Explore and Google Scholar. AMR-producing companies were found using a basic Google search. The starting keywords that were used were “AMR” and “warehouse”. To narrow down the search results other keywords were used such as \textit{“scheduling,”  "decentralized,” “AMR localization,” “fleet size,” “path planning,” “battery management,” [“AMR/AGV” + “hardware”], “zoning,” “warehouse layout,” and [“AMR/AGV” + ”smart manufacturing”]}. Combinations of the above keywords were also used to narrow down the search results. Papers were filtered from 2013 to 2024 with a few references to older papers. Table~\ref{tab1} summarizes the references used in this literature review.

This paper is organized as follows: Section 2 covers the recent AMR technological and software developments, Section 3 goes over robotic control through localization, path planning, and artificial intelligence (AI), Section 4 covers AMR system control by resource management, scheduling, human-robot picking methods, and warehouse flow and layout design, Section 5 goes over future research, and Section 6 is the conclusion.

\section{Hardware}\label{sec2}

\subsection{Sensors}\label{sec3}

Typical AMRs are equipped with a wide variety of 2D and 3D cameras, accelerometers, gyroscopes, and Light Detection and Ranging (LiDAR). The data from these sensors are then brought together in a technique called sensor fusion to generate a map that the AMR can use to locate itself~\cite{de_silva_robust_2018}. These sensors have become popular due to their speedy rendering and positional accuracy. Some popular 2D and 3D cameras are the Balser Dart series~\cite{ag_basler_nodate}, Zivid M60~\cite{zivid_zivid_nodate}, and Nerian Ruby~\cite{noauthor_nerian_nodate}.

One of the most common sensors used in industry is LiDAR. The sensor works by sending out a known waveform into a scene, it is then reflected off of an object, and the LiDAR receiver captures a portion of the waveform and estimates the time it took, which distance is then derived from~\cite{bastos_overview_2021}. The sensors are very accurate and can have a working range of .05m to 25m~\cite{noauthor_2d_nodate}. Some LiDAR sensors rely on artificial landmarks, like reflective surfaces, already placed in the warehouse. These landmarks could become damaged, which could comprise the AMR's localization properties. 

\begin{figure}[t]
\centering
\includegraphics[width=2.5in]{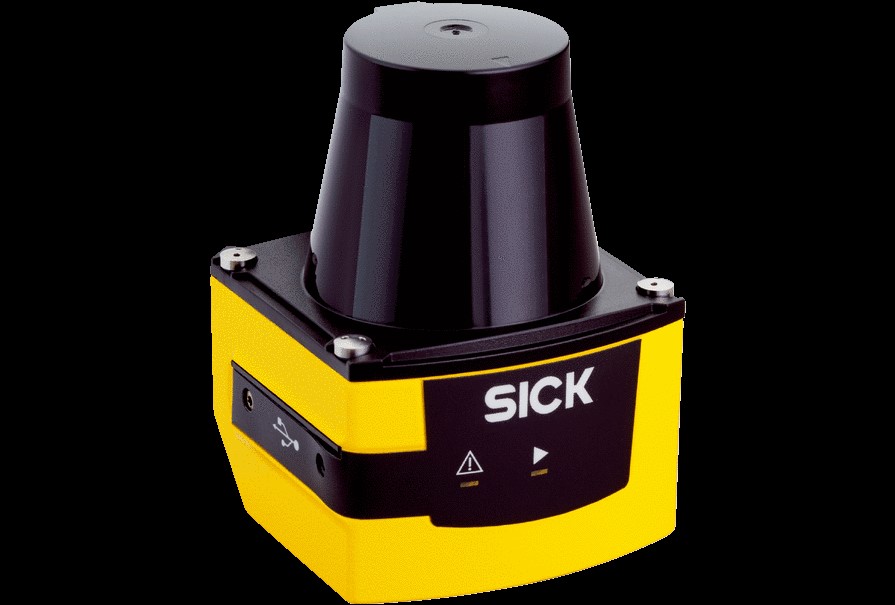}
\caption{Sick TiM-S Safety Laser Scanner~\cite{noauthor_safety_nodate-1}}
\label{Fig: 2}
\end{figure}

Table~\ref{tab2} describes a few AMRs used in the industry today. It shows the hardware that AMRs use as well as describes a few safety features. The SICK laser safety scanner is an example of an AMR safety feature [see Fig.~\ref{Fig: 2}]. The sensors have a scanning angle of 270° and a response time of 67ms~\cite{noauthor_safety_nodate-1}. Scanners like these are often combined with LiDAR to create an all-around safe robot. 

\begin{table}[h]
\caption{AMR Robots}
\label{tab2}
\resizebox{\textwidth}{!}{\begin{minipage}{\textwidth}
\begin{tabular}{@{}ccccccccc@{}}
\toprule
 & \begin{tabular}[c]{@{}c@{}}ADDVERB\\   - Dynamo Series~\cite{noauthor_dynamo_nodate}\end{tabular} & \begin{tabular}[c]{@{}c@{}}MiR - MiR\\   Series~\cite{noauthor_mobile_nodate}\end{tabular} & \begin{tabular}[c]{@{}c@{}}Forward X - Max\\   Series~\cite{noauthor_max_nodate}\end{tabular} & Locus Origin~\cite{noauthor_locus_nodate} & Locus Max~\cite{noauthor_locusmax_nodate} & \begin{tabular}[c]{@{}c@{}}Fetch - \\ Roller Top~\cite{noauthor_rollertop_nodate}\end{tabular} & Matthews AMR~\cite{noauthor_autonomous_nodate} & Bastian - ML2~\cite{noauthor_ml2_nodate} \\ \midrule
Payload(kg) & 100-1000 & 100 - 1350 & 600 - 1500 & 36 & 1361 & 80 & 70 & 200 \\
Speed(m/s) & 2 & 1.5 - 1.5 & - & - & - & 1.5 & 1.8 & 1.8 \\
Run Time(hr) & 3.5-4 & 9.5 - 10 & 8 & 14 & 8-10 & 9 & 6-8 & - \\
Accessorize & Yes & Yes & Yes & \begin{tabular}[c]{@{}c@{}}Multiple\\   config.\end{tabular} & No & No & Yes & Yes \\
\multirow{3}{*}{Features} & \begin{tabular}[c]{@{}c@{}}2D LiDAR, 3D \\ Depth Cameras, \\ Sensors for lower \\ ground obstacle\\   detection\end{tabular} & \begin{tabular}[c]{@{}c@{}}2x SICK laser \\ safety scanners, \\ 2x 3D Depth Cameras, \\ proximity sensors\end{tabular} & \begin{tabular}[c]{@{}c@{}}2x LiDAR sensors, \\ 2x 2D UWA Cameras,\\  2x 3D Depth Cameras, \\ odometer, IMU\end{tabular} & \begin{tabular}[c]{@{}c@{}}8 Sensors\\  and Cameras\end{tabular} & \begin{tabular}[c]{@{}c@{}}x2\\   LiDAR \\ Sensors\end{tabular} & \begin{tabular}[c]{@{}c@{}}2D Laser Sensor,\\   x2 3D Camera,\end{tabular} & \begin{tabular}[c]{@{}c@{}}LiDAR, x2 safety\\   laser scanners\end{tabular} & \begin{tabular}[c]{@{}c@{}}LiDAR, CAT 3\\   Safety system\end{tabular} \\
 & \begin{tabular}[c]{@{}c@{}}Positioning \\ Accuracy(mm): +-20\end{tabular} & \begin{tabular}[c]{@{}c@{}}Positioning\\   Accuracy(mm): +-11\end{tabular} & \begin{tabular}[c]{@{}c@{}}Positioning:\\   Laser SLAM/\\ Visual Tag/\\ Visual Semantics\end{tabular} & \begin{tabular}[c]{@{}c@{}}Tablet\\   based UI, \\ Integrated scanner\end{tabular} & \begin{tabular}[c]{@{}c@{}}Autonomous\\   charging during \\ opportune times\end{tabular} & \begin{tabular}[c]{@{}c@{}}Adjustable top\\   deck height \\ to reach conveyors\end{tabular} & \begin{tabular}[c]{@{}c@{}}Can handle\\   multiple totes and \\ can include a roller \\ conveyor attachment\end{tabular} & \begin{tabular}[c]{@{}c@{}}Small size but\\   many different \\ configurations\end{tabular} \\
 & \begin{tabular}[c]{@{}c@{}}Fast Charging: \\ 20-80\% in 8min\end{tabular} & \begin{tabular}[c]{@{}c@{}}Charging:\\   1.5 hrs – 50min \\ for full charge, \\ 1:6 - 1:12 charging to \\ runtime ratio\end{tabular} & \begin{tabular}[c]{@{}c@{}}Navigation:\\   Natural/\\ Road Network/\\ Hybrid/Follow\end{tabular} & \begin{tabular}[c]{@{}c@{}}Charge\\   Time: 50min\end{tabular} & \begin{tabular}[c]{@{}c@{}}Charge\\   Time: 90min\end{tabular} & \begin{tabular}[c]{@{}c@{}}Charge time: \\ 3hrs for 90\%\end{tabular} & \begin{tabular}[c]{@{}c@{}}Navigation:\\   Natural feature \\ navigation from \\ safety laser \\ scanner input\end{tabular} & \begin{tabular}[c]{@{}c@{}}Navigation:\\   Natural feature\end{tabular} \\ \bottomrule
    \end{tabular}
\end{minipage}}
\end{table}

\begin{figure}[b]
\centering
\includegraphics[width=2.5in]{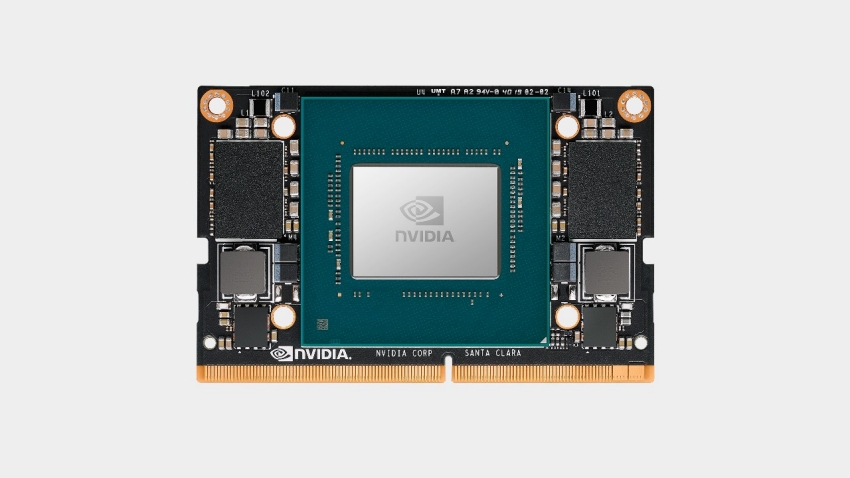}
\caption{NVIDIA Jetson Xavier NX Series\cite{noauthor_nvidia_nodate}}
\label{Fig: 3}
\end{figure}

\subsection{Processors}\label{subsec2}

In order to handle a dynamic work environment, AMRs need to be able to take in data and process it in real-time. Coupled with the advancement of AI, the computational requirements have significantly increased~\cite{gao_performance_2021}. However with the onset of new AI-specific CPUs, GPUs, and AI acceleration units like that of Jetson Xavier NX 16GB [see Fig.~\ref{Fig: 3}], Luxonis DepthAI~\cite{noauthor_depthai_nodate}, and Google’s Coral Accelerator Module~\cite{noauthor_accelerator_nodate}, AI computing has become more reasonable especially for edge computing~\cite{sipola_artificial_2022}. These chips have also become essential to scheduling, especially in decentralized algorithms where each robot must make decisions and communicate messages in a short time frame.

\subsection{Batteries}\label{subsec2}

Batteries that have a high capacity and faster charging have made a significant impact on the robotics industry with most commercially available AMRs using lithium-ion batteries ~\cite{mcnulty_review_2022},~\cite{pistoia_lithium-ion_2013}. Charge times like that of the MiR series of robots have a 1:6 charging to run-time ratio and can have a full charge in 50 minutes~\cite{noauthor_mobile_nodate}. Charging mechanisms have also changed with the onset of wireless power transfer, allowing them to charge without a conventional connector~\cite{huang_modular_2019}. With the onset of these batteries, resource management for battery consumption has become less of a hindrance to warehouse use of AMRs. However, for 24-hour operations, this still has to be taken into consideration. It is important to note that the environmental impact of lithium-ion batteries is still being debated. Currently, there is no method for recycling lithium-ion batteries besides  incineration, which only extracts material that cannot be used to make more batteries~\cite{romare_life_nodate,peters_environmental_2017}.

\section{Robotic Control}\label{sec3}
\subsection{Localization}\label{sec3}

\begin{figure}[t]
\centering
\includegraphics[width=3.0in]{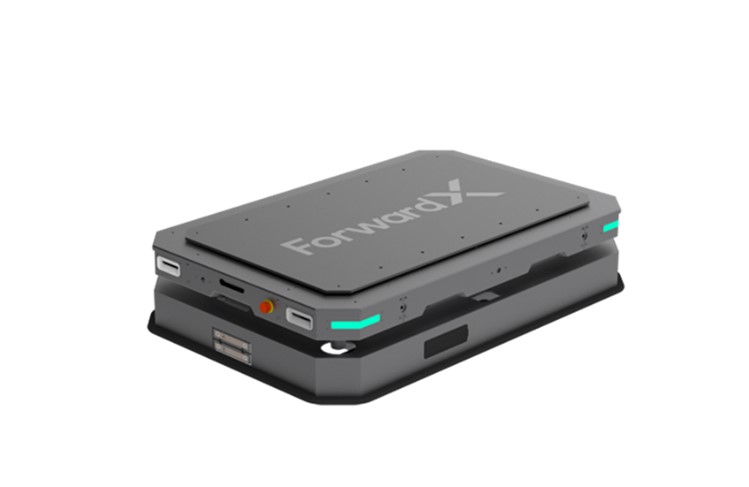}
\caption{FowardX Max series of robots~\cite{noauthor_max_nodate}}
\label{Fig: 4}
\end{figure}

 Many of these robots are equipped with an array of sensors that help locate the robot such as 3D cameras, LiDAR, and encoders. Methods for localization vary from company to company but a common method is SLAM~\cite{singandhupe_review_2019}. SLAM navigation uses features within the environment to build a map. The software gathers information from where the robot is and where it has been to build a probability distribution of all of the likely robot locations~\cite{bloss_simultaneous_2008}. Another common localization method for warehouses is hybrid navigation. This method uses sensor fusion from devices that can locate the robot in large and medium ranges like global positioning system (GPS) GPS (for outdoor environments) or ultrasonic sensors. Then when the robot needs to be located precisely, a short-range sensor (like a magnetic strip) is used to position the robot~\cite{shentu_hybrid_2022}. An example would be the Max series of robots from ForwardX  Fig.~\ref{Fig: 4}. These AMRs have the capability to use SLAM, visual tag, or visual semantics for positioning~\cite{noauthor_flex_nodate}.

\subsection{Artificial Intelligence (AI)}\label{sec3}

AMRs are unique in that they operate in a dynamic environment. When there is an obstacle, the robot needs to be able to autonomously avoid or maneuver itself around the obstruction to continue its mission. Classification of obstacles is commonly done by using a visual system and machine learning. Neural networks, fuzzy logic, or recently deep reinforcement learning are then used to plan a path around the obstacle~\cite{almasri_sensor_2016, yao_multi-robot_2020}. Sensor fusion is also used when certain sensors become unreliable such as GPS when in a covered environment~\cite{gibb_multi-functional_2017}. Without these techniques, AMRs would react to different obstacles in the same manor which becomes impractical when obstacles are consistently changing and moving~\cite{fragapane_planning_2021}. This field of research is very active and continues to improve.

\subsection{Path Planning}\label{sec3}
Path planning is used to plan a trajectory from the current position of the robot to the target using the extracted information from the environment. This problem is typically approached either with a static or dynamic environment. In the warehouse, there are many moving objects like personnel, forklifts, or other autonomous vehicles~\cite{campbell_path_2020}.

For obstacle avoidance, bug algorithms are a common approach to path planning. In general, once an obstacle is detected, the robot will move around the object until it finds the path back to the target objective. Many variations of the bug algorithm exist like that of TangentBug, which finds the shortest path based on a local tangent graph from range data~\cite{kamon_new_1996}. Vector field histograms (VFH) are also a common obstacle avoidance algorithm. In this, a robot detects an obstacle and builds a cell histogram based on the range data. That cell histogram is then converted to a polar histogram where valleys are detected and a path is chosen while taking into consideration the size of the robot~\cite{borenstein_vector_1991}. A variation of the VFH is the VFH* algorithm which uses A* search to build a tree of candidate directions~\cite{ulrich_vfhsup_2000}. Another common approach are artificial potential fields~\cite{woods_novel_2019}. In this, calculated forces are generated for the target and obstacles. When the vehicle moves closer to a obstacle, a repulsive force is acted upon the vehicle avoiding the obstacle while an attractive force from the target keeps the vehicle moving towards the goal. Dang and La~\cite{dang_formation_2019}, extend the artificial potential field by incorporating a rotational force field and a repulsive force field. This new combination helps move a vehicles around complex obstacle shapes were it would otherwise get stuck. 

Heuristic algorithms for path planning have recently been a popular research subject. Examples include neural networks, fuzzy logic, genetic algorithms, and particle swarm optimization. Genetic algorithms~\cite{keith_Dynamic_2024} iterate through potential solutions and find the best selection. This selection is then used to build more solutions and eventually find the optimal path taking advantage of operators like mutation, crossover, and selection~\cite{injarapu_survey_2017}. A hybrid adaptive genetic algorithm is proposed in Zhou \emph{et al.}~\cite{zhou_agv_2022}. The paper addresses the issue of preventing a local optimum by initially performing genetic operations with fixed parameters and then applying adaptive genetic operations. The algorithm adjusts the crossover and mutation probability in sinusoidal adaptive transformation. Fuzzy logic uses the concept of True and False (1 or 0) but extends it to include more freedom between states. First introduced in 1965 by Zadeh~\cite{zadeh_fuzzy_1965}, fuzzy logic algorithms have now advanced to include hybrid path planning like that of Hassanzadeh and Sadigh.~\cite{hassanzadeh_path_2009} or using fuzzy logic with ant colony optimization Song \emph{et al.}~\cite{song_dynamic_2020}. In Song \emph{et al.}, the robots were able to find an optimal path by not only considering the shortest length but factors like the least number of intersections, least traffic, and safety. In another paper by Young and La~\cite{young_consensus_2020}, they combine consensus, cooperative learning, and flocking. In this, a multi agent system is used to first group together the many different actors. Then the algorithm is trained via reinforcement learning to avoid obstacles. Lastly, to figure out where a moving obstacle is coming from, a consensus algorithm~\cite{olfati-saber_consensus_2007} is used. This allows the multi agent system to move as group to avoid oncoming obstacles.

Often, path planning for a single robot is not the most optimal, especially in a warehouse setting where there could be hundreds of robots. Factors like traffic congestion can slow the entire system if not considered. A genetic multi-robot path planning (GMPP) algorithm was proposed in Fan \emph{et al.}~\cite{fan_research_2021}. To avoid collisions with obstacles or other robots the authors integrated a collision detection and elimination mechanism. The collision mechanism calculates where robots will collide then a central planner decides which robots have priority by a random dynamic priority policy. In a recent paper Pradhan \emph{et al.}~\cite{pradhan_novel_2020}, the authors were able to coordinate multi-robot navigation by having a particle swarm optimization algorithm train a feed-forward neural network. They simulated the results in a dynamic environment and compared them to a traditional potential field method~\cite{woods_novel_2019}. A hybrid approach using a convolutional neural network and a graph neural network was proposed in Li \emph{et al.}~\cite{li_graph_2020}. Their approach focuses on decentralizing the multi-robot path planning problem. In Chen \emph{et al.}~\cite{chen_decentralized_2017}, the authors propose a decentralized multi-agent collision avoidance algorithm based on deep reinforcement learning. The approach takes online predicting data for interaction patterns and offloads it to an offline learning procedure. Through this, a robot can predict the behavior of its neighbors even when those neighbors are non-communicative.

\section{System Control}\label{sec4}
\subsection{Resource Management}\label{sec4}
\subsubsection{Battery Use}\label{sec4}
In a warehouse environment, demand often fluctuates. Sometimes orders will surge and strain the available resources within the warehouse. Often bottlenecks are created when there are not enough staff to handle these fluctuations. This can lead to increased lead time for customers, which is undesirable in a very competitive market where on-time delivery is a selling point~\cite{ladel_common_2022}. For a multi-robot system, this must be taken into consideration. Companies like that of ADDVERB~\cite{noauthor_dynamo_nodate} and their AMR called Dynamo, take advantage of lows in demand by using that time to send robots to charge. This helps maximize the robot's utilization and energy management. 

\begin{figure*}[t]
    \centering
    \includegraphics[width=4.0in]{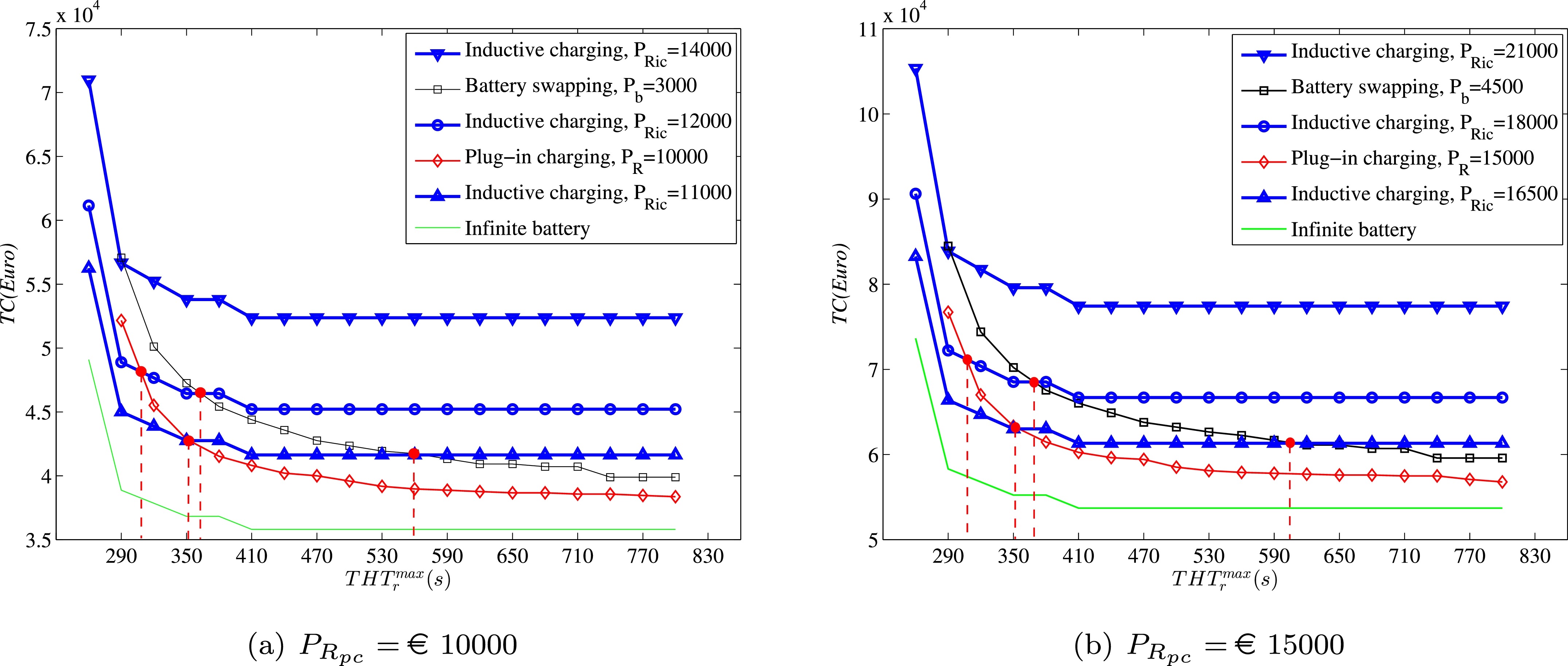}
    \caption{Robot battery cost comparison. $TC(Euro)$: Total cost, $THT_r^{max}(s)$: Required system throughput time, $P_{R_{pc}}$: Plug-in charge robot price~\cite{zou_evaluating_2018}}
    \label{Fig: 5}
\end{figure*}

Traditionally, battery management is usually handled by a rule-based policy. Whenever the battery is low the robot will head to the charging station regardless of where the robot is, demand, or current schedule. This can lead to issues where too many robots charge while there is a surge in demand. Zou \emph{et al.}~\cite{zou_evaluating_2018} compared battery swapping, inductive, and plug-in charging techniques. They showed that inductive charging has the best throughput time
and that battery swapping outperforms plug-in charging. Fig.~\ref{Fig: 5} shows the comparison of the three strategies at different robot price points. Mu \emph{et al.}~\cite{mu_battery_2022} propose a method that models battery management into a Markov Decision Process and then uses a deep reinforcement learning algorithm to solve it. Their results show that their method outperforms standard rule-based strategies by 5\% in fulfillment rate and has significantly fewer backlogged orders after a long period of time. In another paper De Ryck \emph{et al.}~\cite{de_ryck_resource_2020}, the authors use a decentralized approach based on an extension of the traveling salesman problem. Their focus was based on when to charge and how long. Their algorithm also allows the robot to be charged along its current operation trajectory. 

\subsubsection{Fleet Size}\label{sec4}
Determining the correct number of AMRs to use on the floor is important to maintain a consistent lead time. Areas on the warehouse floor where there is a lot of demand can turn into a bottleneck if there are many robots or personnel in the area. Likewise, if there are too few robots, shipments will be paused while orders are being fulfilled. An example of how companies handle this problem is the MiRFleet Software from MiR~\cite{noauthor_centralized_nodate}. This software can handle up to 100 robots with different modules and can coordinate traffic control in critical zones. It allows the user with a basic level of programming to help manage the fleet's priorities. Once the setup and programming are completed, it carries out operations based on the position and availability of the robot.

\begin{figure}[b]
    \centering
    \includegraphics[width=2.5in]{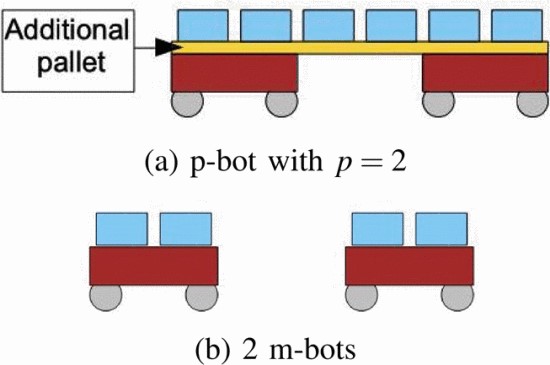}
    \caption{Robot cooperation with p and m-bots~\cite{chaikovskaia_sizing_2021}}
    \label{Fig: 6}
\end{figure}

In literature, the fleet size problem can be solved by queuing theory, linear and integer programming, or discrete and continuous event simulation~\cite{choobineh_fleet_2012}. In a paper by Chaikovskaia \emph{et al.}~\cite{chaikovskaia_sizing_2021}, the authors, using integer linear programming,  considered a fleet under a homogeneous system with robots that can cooperate with each other to carry a load. The paper uses different scenarios where cooperation is inefficient vs. when it is optimal when carrying a load. Fig.~\ref{Fig: 6} shows an example of how p and m-bots work together to carry a load, the p-bot being composed of two m-bots. A modified memetic particle swarm optimization (MMPSO) algorithm was proposed in Chawla \emph{et al.}~\cite{chawla_material_2019}. They first estimated the fleet size by a analytical model and then optimized it using a MMPSO algorithm. They tested with three different floor layouts and compared MMPSO to the analytical model. Vivaldini \emph{et al.}~\cite{vivaldini_integrated_2016}, created a task assignment module that estimates the number of AGVs based on the ratio of defined execution time and total time spent routing. Most articles deal with homogeneous systems with a fleet of the same robot. Rjeb \emph{et al.}~\cite{rjeb_sizing_2021} proposed a method to tackle fleet sizing of a heterogeneous system by an integer linear program. In a 2024 paper by Maurizio \emph{et al.}~\cite{boccia_exact_2024}, the authors propose a genetic algorithm and a mixed-integer linear program to tackle a multi-objective problem consisting of figuring out the best schedule for minimizing makespan and the number of AGVs that are under battery constraints. They compare their results to actual data that was used in a manufacturing facility. The genetic algorithm was shown to outperform the solution that the manufacturing facility was currently implementing. 

\begin{figure*}[h]
    \centering
    \includegraphics[width=\textwidth]{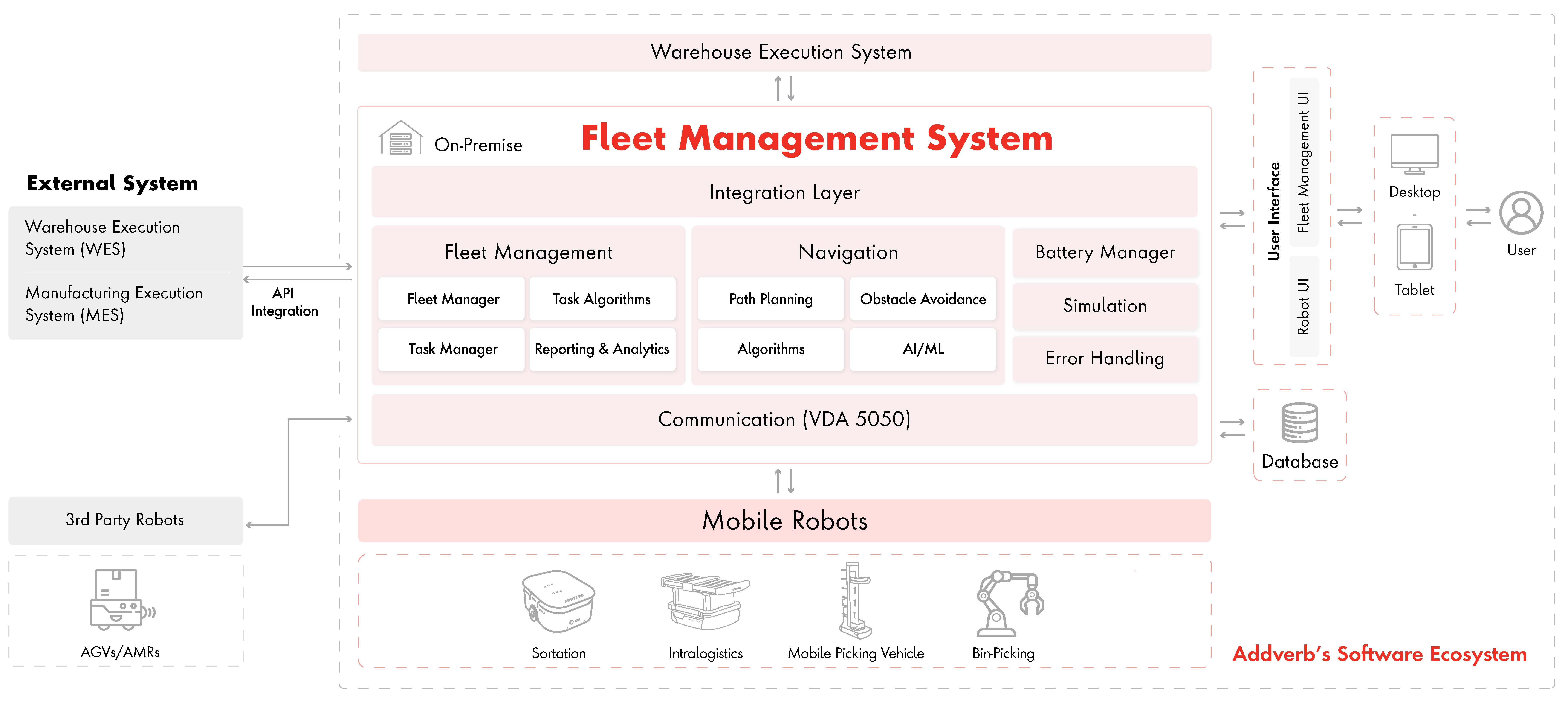}%
    \caption{Fleet management system from ADDVERB~\cite{noauthor_fleet_nodate}}
    \label{Fig: 7}
\end{figure*}

\subsection{Scheduling}\label{sec4}

In academia, a large amount of research has gone into the decision-making process in scheduling orders. Traditionally, these systems are centralized and use hierarchical control. However, with the advancement of computational power, AI has played more of a role in task scheduling in a manufacturing environment~\cite{fragapane_planning_2021}. In industry, scheduling is handled by a warehouse management system (WMS). This software allows administrators to control the daily warehouse operations from the time goods enter to when they move out~\cite{noauthor_what_nodate}. A common example of a WMS software is SAP~\cite{noauthor_sap_nodate}. One of the benefits of using a WMS is that work orders are only given when there is enough inventory to fulfill them. Often this software controls inventory at a high level leaving the planning of how orders are physically picked to lower-level software. ADDVERB Fleet Management System is an example of this lower-level control~\cite{noauthor_fleet_nodate}. It handles the task allocation along with the navigation of each robot. Fig.~\ref{Fig: 7} shows what this software controls and how the user and robot integrate with each other. The traditional approach to scheduling handles task assignment as a path planning problem, assuming the AGV/AMR plans the path from the current location to the target. These algorithms are limited by their flexibility since they do not consider a dynamic work environment.  

Scheduling problems are considered to be NP-hard problems~\cite{anghinolfi_bi-objective_2021}. Due to the complexity, they are normally solved through heuristic algorithms like genetic algorithms, particle swarm optimization, or reinforcement learning. Yokota~\cite{yokota_min-max-strategy-based_2019} proposed an algorithm that uses a min-max strategy to coordinate between robots and humans. The algorithm centers around the co-involvement of robots and humans with the humans picking an item off the shelf and placing it onto an AGV. Each AGV is equipped with a balanced workload based on minimizing the total travel path. Then a heuristic algorithm, with two different rules, is then used to match a picker with a robot. In Yu \emph{et al.}~\cite{yu_multi-load_2021}, a algorithm is proposed using a two-stage heuristic algorithm. The first stage uses clustering to divide packages based on similarity and group them together with the idea that a multi-load AGV can sort the packages of one group in one operation. Then in the second stage, a load-balancing scheduling algorithm assigns and sorts each of the groups to the AGVs. A hierarchical soft actor-critic algorithm is proposed in Tang \emph{et al.}~\cite{tang_novel_2021}. Hierarchical reinforcement learning algorithms attempt to reduce the computational complexity of a dynamic environment by layering learning strategies. Soft Actor-Critic (SAC) algorithms are more practical in robot control since they are able to solve both discrete and continuous problems. The algorithm centrally assign tasks to the AGVs then the robots collaborate in planning the actual path. In another paper Li and Wu~\cite{li_research_2022}, proposed a Genetic Algorithm Considering Genome, which takes into account dynamic task scheduling as well as determining the charging schedule. It is able to iterate through a genome, which is composed of several neighborhood solutions. Their algorithm can quickly find a local optimal solution and, in a later stage, explore the neighborhood of the the local optimal solution using internal cross and mutation. In a 2024 paper, Li and Huang~\cite{li_efficient_2024} propose a scheduling algorithm for heterogeneous AGVs. This allows for the use of several different kinds of AGVs performing different tasks. They were able to implement several different algorithms including Fist Come First Assign, Heterogeneous Task Assignment, and Optimization Assignment and compare them to past works. 

Few studies have been conducted that offer a way to decentralize scheduling. Hierarchical control often finds the optimal solution but suffers from a lack of robustness or flexibility. If the central control unit goes down the entire system cannot operate. Decentralized control has the advantage of scalability and the ability to continue operation even if one unit fails. Basile \emph{et al.}~\cite{basile_auction-based_2017}, used a version of an auction-based approach called the sealed-bid method in which an agent cannot see other agents’ bids. Any robot can play the role of auctioneer and takes bids in from other robots acting as an agent. The auctioneer selects the best bid based on the time it takes to complete a mission. In another paper by Warita and Fujita~\cite{warita_online_2024}, the authors set up a multi-agent decentralized algorithm using a fully decoupled upper confidence tree (FDUCT). The idea being that each AGV will use FDUCT to plan its own actions while predicting the actions of other agents. They also implement an item-exchange strategy to help balance the load of each AGV and avoid idle time. 

\subsection{Human-Robot Picking Methods}
\subsubsection{Picking Methods}
\begin{figure}[t]
    \centering
    \includegraphics[width=3.4in]{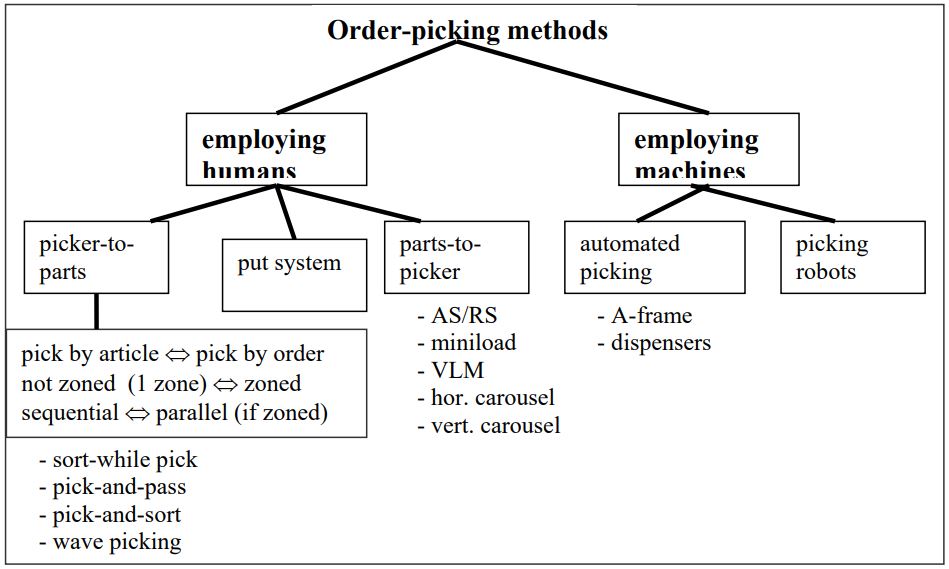}
    \caption{Traditional picking methods~\cite{de_koster_design_2007}}
    \label{Fig: 8}
\end{figure}

Picking is defined as the process of scheduling customers' orders, assigning stock, releasing orders, picking items off of storage locations, and disposal. The majority of warehouses that employ humans deploy a picker-to-parts system, which involves the picker walking or driving to pick items. More automated warehouses will use a parts-to-picker method where shelving robot units or cranes move products to the picker[see Fig:~\ref{Fig: 8}]. Within picker-to-parts systems, there are several variants like batch picking or zoning. Batch picking has the picker pick multiple orders at once then sorted in a later sequence. Zoning is a system in which the warehouse floor is divided into areas where humans pick items that are only in their designated area~\cite{de_koster_design_2007, boysen_parts--picker_2017}.

\subsubsection{Human-Robot Collaboration}
In AMR-assisted picking, the laborious task of traveling with a picked item is reduced by the use of a robot. In a paper by Srinivas and Yu~\cite{srinivas_collaborative_2022}, a collaborative human-robot order-picking system (CHR-OPS) was presented with the goal of minimizing the tardiness in their batch order system. Three sub-problems were made to optimize; the number of items picked in one tour, batch assignment and sequencing, and picker-robot routing. Their results emphasized the impact of AMR cart capacity, AMR speed, and human-robot team composition. In another picker-to-parts system, Zulj Ivan \emph{et al.}~\cite{zulj_order_2022} used zoning to have humans pick batched items and drop them off in pick location for that zone. A robot would then collect and transport the batched items to a depot where they will be sorted. Their approach focuses on minimizing the tardiness of all customer orders. The algorithm used a two-stage heuristic of adaptive large neighborhood search (ALNS) for batching and NEH heuristic for the sequencing the batches. Zhao \emph{et al.}~\cite{zhao_order_2024} proposed a human-robot collaboration algorithm in a parts-to-picker system. Their algorithm uses an adaptive large neighborhood search method that is embedded in a tabu search algorithm. They compare their algorithm with others and show it effectiveness at lowering makespan. 

\subsection{Warehouse Flow and Layout Design}

\begin{figure}[t]
    \centering
    \includegraphics[width=1.8in]{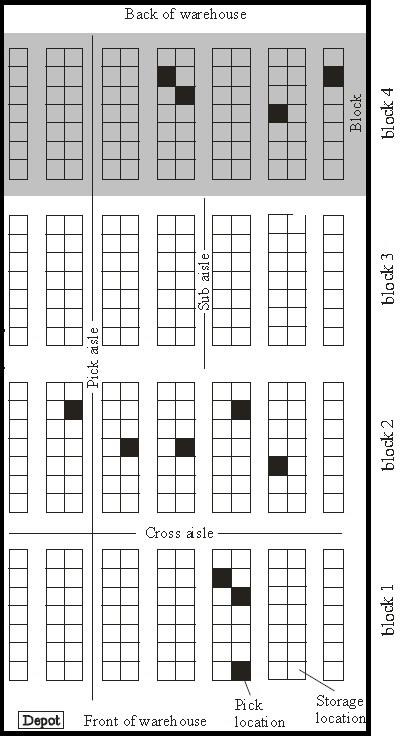}
    \caption{Conventional warehouse layout~\cite{roodbergen_designing_2008}}
    \label{Fig: 9}
\end{figure}

\begin{figure}[t]
    \centering
    \includegraphics[width=2.8in]{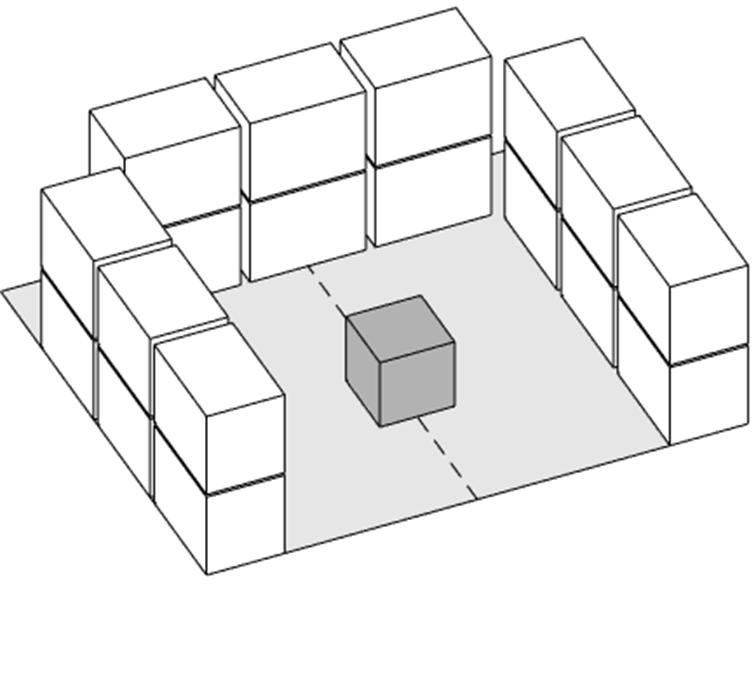}
    \caption{Non-conventional U-shaped warehouse layout~\cite{diefenbach_ergonomic_2019}}
    \label{Fig: 10}
\end{figure}

Traditionally, warehouse layouts can be classified as a conventional, non-conventional, and general warehouse. Conventional warehouses are those with a rectangular shape and parallel aisles that are perpendicular to straight cross aisles. Layouts with two cross aisles are considered as single-block warehouses. Fig.~\ref{Fig: 9} shows a two-block layout with three cross aisles. Non-conventional warehouses arrange their aisles in a manner that allows easier access to certain areas of the warehouse [see Fig.~\ref{Fig: 10}]. For example, fishbone warehouse layouts offer a scheme that is shown to reduce travel distance by 10-15\% when compared to conventional warehouses~\cite{wan_integrating_2022}. General warehouses do not make any assumptions about the aisle locations. Rather, they are modeled after general distance matrices~\cite{masae_order_2020}. The layout also defines aisle characteristics. For a multi-robot system, wide aisles are a benefit since they allow room for robots to move around. However, narrow aisles reduce the distance traveled for items that are picked on either side of the aisle. Items that are picked low to the ground and do require any vertical tools to grab are considered low-level~\cite{scholz_order_2017,cano_solving_2023,manzini_warehousing_2012}

\begin{figure}[t]
    \centering
    \includegraphics[width=3.3in]{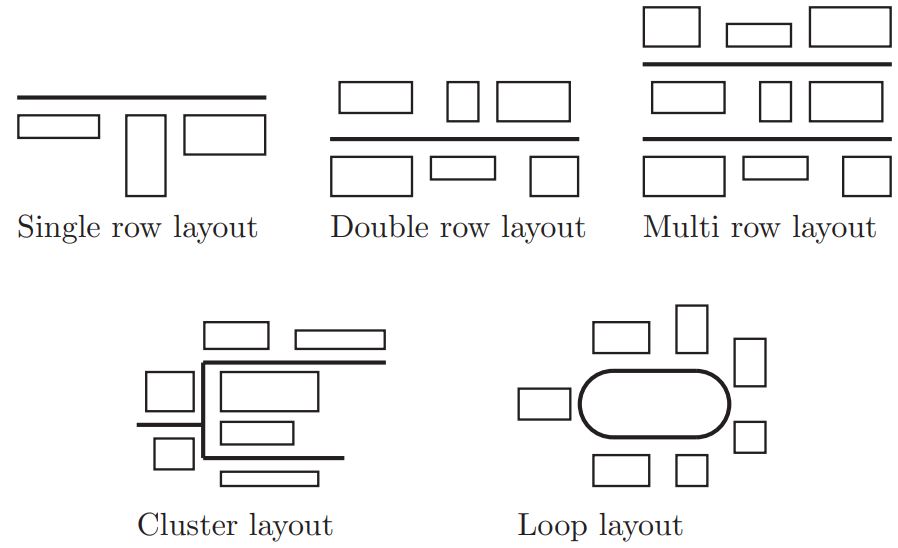}
    \caption{Classifications of row layout ~\cite{keller_single_2015}}
    \label{Fig: 11}
\end{figure}

In a paper by Hamzeei \emph{et al.}~\cite{hamzeei_exact_2013}, they investigated the bidirectional topology of a block layout design of a warehouse floor and the locations of delivery and pickup stations. They developed two algorithms; one used a cutting-plane algorithm, and the other used simulated annealing to solve the problem heuristically. Li and Li ~\cite{li_bi-objective_2023} looked at optimizing a multi-row layout of a machining workshop while considering AGV path flow. The multi-row layout consists of robots covering an area that has multiple rows and can be picked on either side [see Fig.~\ref{Fig: 11}]. The proposed algorithm uses a hybrid method between non-dominated sorting genetic algorithm-II and tabu search. By focusing on the AGV path they were able to lower the material handling cost of the facility~\cite{keller_single_2015}. In a 2023 paper by Zhang \emph{et al.}~\cite{zhang_multi-robot_2023}, the authors consider the case of a fully automated warehouse with no human personal. Since robots do not need a structured layout to locate parts, the authors make the case that an optimized layout, that does not resemble a human-designed layout, can be used to increase throughput. The layouts that are generated from their algorithm are non-traditional and do not have column-row aisles in a typical warehouse layout. To determine the storage location of items Bao \emph{et al.}~\cite{bao_storage_2019}, classified items based on their profit or throughput rate. Products that are labeled as “A” would be placed closest to the output gate to reduce travel time.

\section{Future Work}
Scheduling has normally been focused on a centralized network of robots. However, this creates a dependency on the central unit and demands one unit to do most of the computational work. Decentralizing this task could prove to be more flexible and scalable. The papers presented~\cite{yokota_min-max-strategy-based_2019, yu_multi-load_2021,tang_novel_2021,li_research_2022} simulate 3 to 10 robots at a time. In real-world applications, warehouses could deploy hundreds~\cite{wessling_chery_2023}. Only a few studies were found that deal with decentralized scheduling ~\cite{basile_auction-based_2017}, ~\cite{gehlhoff_agent-based_2020}.

Warehouse zoning is important for dividing the workload on the warehouse floor. For AMR networks this means faster response time and less likelihood of congestion. Most of the studies presented~\cite{hamzeei_exact_2013,li_bi-objective_2023,keller_single_2015,chen_using_2018,ho_zone_2009,saylam_minmax_2023,li_multi-agvs_2019,bao_storage_2019,yuan_bot--time_2017,tang_research_2021,correia_implementing_2020,ilic_analysis_1994} focus on AGV vehicles. One of the downsides of using AGVs, are the pre-planned routes they must follow. AMRs do have to deal with this issue since they can find alternate routes around obstacles or other robots without the need of a user designing an optimal path. More research needs to be done into modeling dynamic zones for AMRs to balance workloads and ensure a rapid response.

\begin{figure}[t]
    \centering
    \includegraphics[width=4.4in]{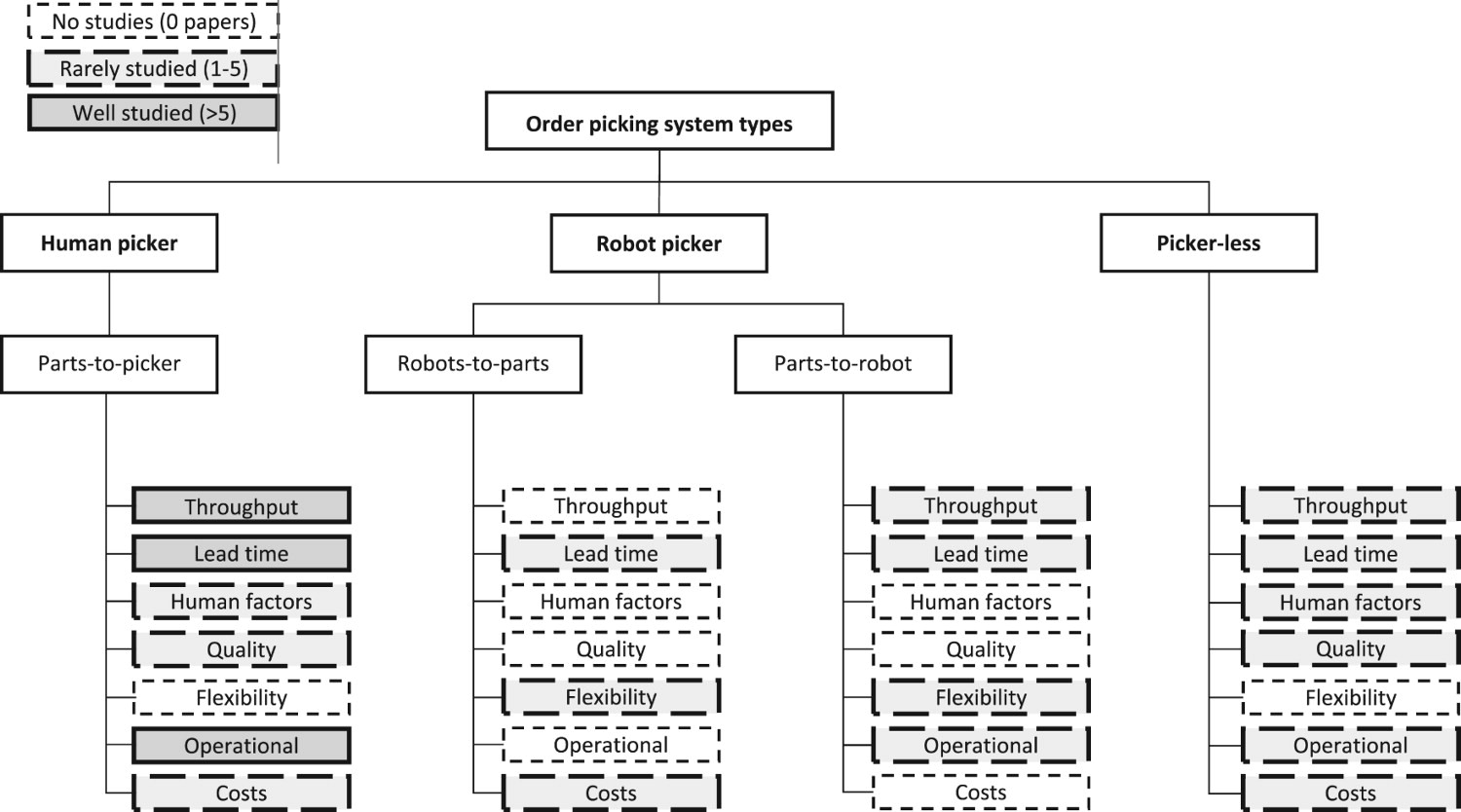}
    \caption{Areas of focus within part-to-picker systems~\cite{jaghbeer_automated_2020}}
    \label{Fig: 12}
\end{figure}

Many papers use performance indicators like lead time, tardiness, distance traveled, and operational efficiency to measure how well their algorithm works when compared to others. However, in human/robot collaborative spaces, human factors are rarely considered. In another literature review that looked at parts-to-picker systems, Jaghbeer \emph{et al.}~\cite{jaghbeer_automated_2020} highlighted that human factors are rarely focused on in picker-less, robot-to-parts, or parts-to-robot OPSs (order picking systems) [see Fig:~\ref{Fig: 12} ]. 

\section{Conclusion}
Autonomous mobile robots (AMRs) have made a significant contribution to manufacturing specifically in warehousing improving performance and productivity. Several technological developments in localization, sensors, and battery management have AMRs more feasible to be deployed. AI techniques have also played an important role in the decision-making of AMRs. Genetic algorithms, deep reinforcement learning, and neural networks have reduced the complexity of scheduling paving the way for decentralization. Different warehouse layouts, picking strategies, and human-robot routing strategies have helped in the development of human-robot collaboration. With this research, AMRs have shared the workload of laborious and repetitive tasks allowing humans to be less fatigued and focused. Though most of the research that has been presented in this paper has focused on warehousing, not enough has been done studying applications in other environments such as outdoor warehousing of docking yards or hazardous areas. It is concluded that the research in this field is growing rapidly and is currently changing the manufacturing industry.

\bibliography{sn-article}


\begin{thebibliography}{107}
\ifx \bisbn   \undefined \def \bisbn  #1{ISBN #1}\fi
\ifx \binits  \undefined \def \binits#1{#1}\fi
\ifx \bauthor  \undefined \def \bauthor#1{#1}\fi
\ifx \batitle  \undefined \def \batitle#1{#1}\fi
\ifx \bjtitle  \undefined \def \bjtitle#1{#1}\fi
\ifx \bvolume  \undefined \def \bvolume#1{\textbf{#1}}\fi
\ifx \byear  \undefined \def \byear#1{#1}\fi
\ifx \bissue  \undefined \def \bissue#1{#1}\fi
\ifx \bfpage  \undefined \def \bfpage#1{#1}\fi
\ifx \blpage  \undefined \def \blpage #1{#1}\fi
\ifx \burl  \undefined \def \burl#1{\textsf{#1}}\fi
\ifx \doiurl  \undefined \def \doiurl#1{\url{https://doi.org/#1}}\fi
\ifx \betal  \undefined \def \betal{\textit{et al.}}\fi
\ifx \binstitute  \undefined \def \binstitute#1{#1}\fi
\ifx \binstitutionaled  \undefined \def \binstitutionaled#1{#1}\fi
\ifx \bctitle  \undefined \def \bctitle#1{#1}\fi
\ifx \beditor  \undefined \def \beditor#1{#1}\fi
\ifx \bpublisher  \undefined \def \bpublisher#1{#1}\fi
\ifx \bbtitle  \undefined \def \bbtitle#1{#1}\fi
\ifx \bedition  \undefined \def \bedition#1{#1}\fi
\ifx \bseriesno  \undefined \def \bseriesno#1{#1}\fi
\ifx \blocation  \undefined \def \blocation#1{#1}\fi
\ifx \bsertitle  \undefined \def \bsertitle#1{#1}\fi
\ifx \bsnm \undefined \def \bsnm#1{#1}\fi
\ifx \bsuffix \undefined \def \bsuffix#1{#1}\fi
\ifx \bparticle \undefined \def \bparticle#1{#1}\fi
\ifx \barticle \undefined \def \barticle#1{#1}\fi
\bibcommenthead
\ifx \bconfdate \undefined \def \bconfdate #1{#1}\fi
\ifx \botherref \undefined \def \botherref #1{#1}\fi
\ifx \url \undefined \def \url#1{\textsf{#1}}\fi
\ifx \bchapter \undefined \def \bchapter#1{#1}\fi
\ifx \bbook \undefined \def \bbook#1{#1}\fi
\ifx \bcomment \undefined \def \bcomment#1{#1}\fi
\ifx \oauthor \undefined \def \oauthor#1{#1}\fi
\ifx \citeauthoryear \undefined \def \citeauthoryear#1{#1}\fi
\ifx \endbibitem  \undefined \def \endbibitem {}\fi
\ifx \bconflocation  \undefined \def \bconflocation#1{#1}\fi
\ifx \arxivurl  \undefined \def \arxivurl#1{\textsf{#1}}\fi
\csname PreBibitemsHook\endcsname

\bibitem[\protect\citeauthoryear{Muller}{}]{muller_automated_1983}
\begin{botherref}
\oauthor{\bsnm{Muller}, \binits{T.}}:
{AUTOMATED} {GUIDED} {VEHICLES}.
Accessed 2023-08-14
\end{botherref}
\endbibitem

\bibitem[\protect\citeauthoryear{Longhan et~al.}{}]{longhan_production_2013}
\begin{botherref}
\oauthor{\bsnm{Longhan}, \binits{Z.}},
\oauthor{\bsnm{Hong}, \binits{L.}},
\oauthor{\bsnm{Shiwei}, \binits{X.}}:
Production process improvement based on value stream mapping for {CY} company.
In: 2013 6th International Conference on Information Management, Innovation Management and Industrial Engineering,
vol. 3,
pp. 226--229.
\doiurl{10.1109/ICIII.2013.6703555} .
{ISSN}: 2155-1472
\end{botherref}
\endbibitem

\bibitem[\protect\citeauthoryear{Banker}{}]{banker_new_nodate}
\begin{botherref}
\oauthor{\bsnm{Banker}, \binits{S.}}:
New Robotic Solutions For The Warehouse.
Section: Transportation - Business.
\url{https://www.forbes.com/sites/stevebanker/2017/03/07/new-robotic-solutions-for-the-warehouse/}
Accessed 2023-08-05
\end{botherref}
\endbibitem

\bibitem[\protect\citeauthoryear{Kamali}{}]{kamali_smart_2019}
\begin{botherref}
\oauthor{\bsnm{Kamali}, \binits{D.A.}}:
Smart warehouse vs. traditional warehouse - review
\textbf{11}(1)
\end{botherref}
\endbibitem

\bibitem[\protect\citeauthoryear{Chain}{}]{chain_how_nodate}
\begin{botherref}
\oauthor{\bsnm{Chain}, \binits{N.W.} \bsuffix{Körber~Supply}}:
How Autonomous Mobile Robots Can Solve Global E-Commerce Fulfillment Problems.
\url{https://www.robotics247.com/article/how_autonomous_mobile_robots_can_solve_global_ecommerce_fulfillment_problems}
Accessed 2023-09-03
\end{botherref}
\endbibitem

\bibitem[\protect\citeauthoryear{Kim}{}]{kim_amazon_nodate}
\begin{botherref}
\oauthor{\bsnm{Kim}, \binits{E.}}:
Amazon Is Now Using a Whole Lot More of the Robots from the Company It Bought for \$775 Million.
\url{https://www.businessinsider.com/amazon-doubled-the-number-of-kiva-robots-2015-10}
Accessed 2023-08-21
\end{botherref}
\endbibitem

\bibitem[\protect\citeauthoryear{De~Silva et~al.}{}]{de_silva_robust_2018}
\begin{botherref}
\oauthor{\bsnm{De~Silva}, \binits{V.}},
\oauthor{\bsnm{Roche}, \binits{J.}},
\oauthor{\bsnm{Kondoz}, \binits{A.}}:
Robust fusion of {LiDAR} and wide-angle camera data for autonomous mobile robots
\textbf{18}(8),
2730
\doiurl{10.3390/s18082730} .
Number: 8 Publisher: Multidisciplinary Digital Publishing Institute.
Accessed 2023-08-07
\end{botherref}
\endbibitem

\bibitem[\protect\citeauthoryear{{AG}}{}]{ag_basler_nodate}
\begin{botherref}
\oauthor{\bsnm{{AG}}, \binits{B.}}:
Basler Dart – Area Scan Cameras.
\url{https://www.baslerweb.com/en/products/cameras/area-scan-cameras/dart/}
Accessed 2023-09-02
\end{botherref}
\endbibitem

\bibitem[\protect\citeauthoryear{Zivid}{}]{zivid_zivid_nodate}
\begin{botherref}
\oauthor{\bsnm{Zivid}}:
Zivid 2 Plus M60.
\url{https://www.zivid.com/zivid-2-plus-m60}
Accessed 2023-09-02
\end{botherref}
\endbibitem

\bibitem[\protect\citeauthoryear{}{}]{noauthor_nerian_nodate}
\begin{botherref}
Nerian Ruby.
\url{https://www.alliedvision.com/en/products/3d-cameras/nerian-ruby/}
Accessed 2023-09-02
\end{botherref}
\endbibitem

\bibitem[\protect\citeauthoryear{Bastos et~al.}{}]{bastos_overview_2021}
\begin{botherref}
\oauthor{\bsnm{Bastos}, \binits{D.}},
\oauthor{\bsnm{Monteiro}, \binits{P.P.}},
\oauthor{\bsnm{Oliveira}, \binits{A.S.R.}},
\oauthor{\bsnm{Drummond}, \binits{M.V.}}:
An overview of {LiDAR} requirements and techniques for autonomous driving.
In: 2021 Telecoms Conference (ConfTELE),
pp. 1--6.
\doiurl{10.1109/Conf℡E50222.2021.9435580}
\end{botherref}
\endbibitem

\bibitem[\protect\citeauthoryear{}{}]{noauthor_2d_nodate}
\begin{botherref}
2D {LiDAR} Sensors {\textbar} {TiM}-P {SICK}.
\url{https://www.sick.com/ag/en/lidar-sensors/2d-lidar-sensors/tim-p/c/g568273}
Accessed 2023-08-14
\end{botherref}
\endbibitem

\bibitem[\protect\citeauthoryear{}{}]{noauthor_safety_nodate-1}
\begin{botherref}
Safety Laser Scanners {\textbar} {TiM}-S {\textbar} {SICK}.
\url{https://www.sick.com/pl/en/safety-laser-scanners/safety-laser-scanners/tim-s/c/g502253}
Accessed 2023-08-07
\end{botherref}
\endbibitem

\bibitem[\protect\citeauthoryear{}{}]{noauthor_dynamo_nodate}
\begin{botherref}
Dynamo Archives.
\url{https://addverb.com/product-category/mobile-robots/product-amr/}
Accessed 2023-08-05
\end{botherref}
\endbibitem

\bibitem[\protect\citeauthoryear{}{}]{noauthor_mobile_nodate}
\begin{botherref}
Mobile {Industrial} {Robots} - {Automate} your internal transportation.
Section: page.
\url{https://mobile-industrial-robots.com/}
Accessed 2024-06-05
\end{botherref}
\endbibitem

\bibitem[\protect\citeauthoryear{}{}]{noauthor_max_nodate}
\begin{botherref}
Max Series {AMRs} Autonomous Mobile Robots Warehouse.
\url{https://www.forwardx.com/max-amr/}
Accessed 2023-08-21
\end{botherref}
\endbibitem

\bibitem[\protect\citeauthoryear{}{}]{noauthor_locus_nodate}
\begin{botherref}
Locus Origin.
\url{https://locusrobotics.com/products/locus-origin/}
Accessed 2023-08-07
\end{botherref}
\endbibitem

\bibitem[\protect\citeauthoryear{}{}]{noauthor_locusmax_nodate}
\begin{botherref}
Locus Max.
\url{https://locusrobotics.com/products/locus-max/}
Accessed 2023-08-21
\end{botherref}
\endbibitem

\bibitem[\protect\citeauthoryear{}{}]{noauthor_rollertop_nodate}
\begin{botherref}
{RollerTop} – Fetch Robotics.
\url{https://fetchrobotics.com/rollertop/}
Accessed 2023-08-08
\end{botherref}
\endbibitem

\bibitem[\protect\citeauthoryear{}{}]{noauthor_autonomous_nodate}
\begin{botherref}
Autonomous Mobile Robots.
\url{https://matthewsautomation.com/autonomous-mobile-robots/}
Accessed 2023-08-10
\end{botherref}
\endbibitem

\bibitem[\protect\citeauthoryear{}{}]{noauthor_ml2_nodate}
\begin{botherref}
{ML}2 {AV} {\textbar} Miniload Autonomous Vehicle {\textbar} Bastian Solutions.
\url{https://www.bastiansolutions.com/solutions/technology/automated-guided-vehicles/ml2/#specifications}
Accessed 2023-08-10
\end{botherref}
\endbibitem

\bibitem[\protect\citeauthoryear{}{}]{noauthor_nvidia_nodate}
\begin{botherref}
{NVIDIA} Jetson Xavier Series.
\url{https://www.nvidia.com/en-us/autonomous-machines/embedded-systems/jetson-xavier-series/}
Accessed 2023-08-08
\end{botherref}
\endbibitem

\bibitem[\protect\citeauthoryear{Gao and Song}{}]{gao_performance_2021}
\begin{botherref}
\oauthor{\bsnm{Gao}, \binits{R.}},
\oauthor{\bsnm{Song}, \binits{M.}}:
Performance comparative analysis of artificial intelligence chip technology.
In: 2021 2nd International Conference on Computer Engineering and Intelligent Control ({ICCEIC}),
pp. 149--153.
\doiurl{10.1109/ICCEIC54227.2021.00037}
\end{botherref}
\endbibitem

\bibitem[\protect\citeauthoryear{}{}]{noauthor_depthai_nodate}
\begin{botherref}
{DepthAI}.
\url{https://www.crowdsupply.com/luxonis/depthai}
Accessed 2023-08-08
\end{botherref}
\endbibitem

\bibitem[\protect\citeauthoryear{}{}]{noauthor_accelerator_nodate}
\begin{botherref}
Accelerator Module Datasheet.
\url{https://coral.ai/docs/module/datasheet/}
Accessed 2023-08-08
\end{botherref}
\endbibitem

\bibitem[\protect\citeauthoryear{Sipola et~al.}{}]{sipola_artificial_2022}
\begin{botherref}
\oauthor{\bsnm{Sipola}, \binits{T.}},
\oauthor{\bsnm{Alatalo}, \binits{J.}},
\oauthor{\bsnm{Kokkonen}, \binits{T.}},
\oauthor{\bsnm{Rantonen}, \binits{M.}}:
Artificial intelligence in the {IoT} era: A review of edge {AI} hardware and software.
In: 2022 31st Conference of Open Innovations Association ({FRUCT}),
pp. 320--331.
\doiurl{10.23919/FRUCT54823.2022.9770931}
\end{botherref}
\endbibitem

\bibitem[\protect\citeauthoryear{{McNulty} et~al.}{}]{mcnulty_review_2022}
\begin{botherref}
\oauthor{\bsnm{{McNulty}}, \binits{D.}},
\oauthor{\bsnm{Hennessy}, \binits{A.}},
\oauthor{\bsnm{Li}, \binits{M.}},
\oauthor{\bsnm{Armstrong}, \binits{E.}},
\oauthor{\bsnm{Ryan}, \binits{K.M.}}:
A review of li-ion batteries for autonomous mobile robots: Perspectives and outlook for the future
\textbf{545},
231943
\doiurl{10.1016/j.jpowsour.2022.231943} .
Accessed 2023-08-08
\end{botherref}
\endbibitem

\bibitem[\protect\citeauthoryear{Pistoia}{}]{pistoia_lithium-ion_2013}
\begin{botherref}
\oauthor{\bsnm{Pistoia}, \binits{G.}}:
Lithium-Ion Batteries: Advances and Applications.
Newnes.
Google-Books-{ID}: {wWciAQAAQBAJ}
\end{botherref}
\endbibitem

\bibitem[\protect\citeauthoryear{Huang et~al.}{}]{huang_modular_2019}
\begin{botherref}
\oauthor{\bsnm{Huang}, \binits{S.-J.}},
\oauthor{\bsnm{Lee}, \binits{T.-S.}},
\oauthor{\bsnm{Li}, \binits{W.-H.}},
\oauthor{\bsnm{Chen}, \binits{R.-Y.}}:
Modular on-road {AGV} wireless charging systems via interoperable power adjustment
\textbf{66}(8),
5918--5928
\doiurl{10.1109/TIE.2018.2873165} .
Conference Name: {IEEE} Transactions on Industrial Electronics
\end{botherref}
\endbibitem

\bibitem[\protect\citeauthoryear{Romare and Dahllöf}{}]{romare_life_nodate}
\begin{botherref}
\oauthor{\bsnm{Romare}, \binits{M.}},
\oauthor{\bsnm{Dahllöf}, \binits{L.}}:
The life cycle energy consumption and greenhouse gas emissions from lithium-ion batteries
\end{botherref}
\endbibitem

\bibitem[\protect\citeauthoryear{Peters et~al.}{}]{peters_environmental_2017}
\begin{botherref}
\oauthor{\bsnm{Peters}, \binits{J.F.}},
\oauthor{\bsnm{Baumann}, \binits{M.}},
\oauthor{\bsnm{Zimmermann}, \binits{B.}},
\oauthor{\bsnm{Braun}, \binits{J.}},
\oauthor{\bsnm{Weil}, \binits{M.}}:
The environmental impact of li-ion batteries and the role of key parameters – a review
\textbf{67},
491--506
\doiurl{10.1016/j.rser.2016.08.039} .
Accessed 2023-08-08
\end{botherref}
\endbibitem

\bibitem[\protect\citeauthoryear{Singandhupe and La}{2019}]{singandhupe_review_2019}
\begin{bchapter}
\bauthor{\bsnm{Singandhupe}, \binits{A.}},
\bauthor{\bsnm{La}, \binits{H.M.}}:
\bctitle{A {Review} of {SLAM} {Techniques} and {Security} in {Autonomous} {Driving}}.
In: \bbtitle{2019 {Third} {IEEE} {International} {Conference} on {Robotic} {Computing} ({IRC})},
pp. \bfpage{602}--\blpage{607}.
\bpublisher{IEEE},
\blocation{Naples, Italy}
(\byear{2019}).
\doiurl{10.1109/IRC.2019.00122} .
\burl{https://ieeexplore.ieee.org/document/8675575/}
Accessed 2024-06-05
\end{bchapter}
\endbibitem

\bibitem[\protect\citeauthoryear{Bloss}{}]{bloss_simultaneous_2008}
\begin{botherref}
\oauthor{\bsnm{Bloss}, \binits{R.}}:
Simultaneous sensing of location and mapping for autonomous robots
\textbf{28}(2),
102--107
\doiurl{10.1108/02602280810856651} .
Publisher: Emerald Group Publishing Limited.
Accessed 2023-08-06
\end{botherref}
\endbibitem

\bibitem[\protect\citeauthoryear{Shentu et~al.}{}]{shentu_hybrid_2022}
\begin{botherref}
\oauthor{\bsnm{Shentu}, \binits{S.}},
\oauthor{\bsnm{Gong}, \binits{Z.}},
\oauthor{\bsnm{Liu}, \binits{X.-J.}},
\oauthor{\bsnm{Liu}, \binits{Q.}},
\oauthor{\bsnm{Xie}, \binits{F.}}:
Hybrid navigation system based autonomous positioning and path planning for mobile robots
\textbf{35}(1),
109
\doiurl{10.1186/s10033-022-00775-4} .
Accessed 2023-08-05
\end{botherref}
\endbibitem

\bibitem[\protect\citeauthoryear{}{}]{noauthor_flex_nodate}
\begin{botherref}
Flex Series {AMRs} Order Picking Robot For Warehouse.
\url{https://www.forwardx.com/flex-amr/}
Accessed 2023-08-05
\end{botherref}
\endbibitem

\bibitem[\protect\citeauthoryear{Almasri et~al.}{}]{almasri_sensor_2016}
\begin{botherref}
\oauthor{\bsnm{Almasri}, \binits{M.}},
\oauthor{\bsnm{Elleithy}, \binits{K.}},
\oauthor{\bsnm{Alajlan}, \binits{A.}}:
Sensor fusion based model for collision free mobile robot navigation
\textbf{16}(1),
24
\doiurl{10.3390/s16010024} .
Number: 1 Publisher: Multidisciplinary Digital Publishing Institute.
Accessed 2023-08-05
\end{botherref}
\endbibitem

\bibitem[\protect\citeauthoryear{Yao et~al.}{}]{yao_multi-robot_2020}
\begin{botherref}
\oauthor{\bsnm{Yao}, \binits{S.}},
\oauthor{\bsnm{Chen}, \binits{G.}},
\oauthor{\bsnm{Pan}, \binits{L.}},
\oauthor{\bsnm{Ma}, \binits{J.}},
\oauthor{\bsnm{Ji}, \binits{J.}},
\oauthor{\bsnm{Chen}, \binits{X.}}:
Multi-robot collision avoidance with map-based deep reinforcement learning.
In: 2020 {IEEE} 32nd International Conference on Tools with Artificial Intelligence ({ICTAI}),
pp. 532--539.
\doiurl{10.1109/ICTAI50040.2020.00088} .
{ISSN}: 2375-0197
\end{botherref}
\endbibitem

\bibitem[\protect\citeauthoryear{Gibb et~al.}{2017}]{gibb_multi-functional_2017}
\begin{bchapter}
\bauthor{\bsnm{Gibb}, \binits{S.}},
\bauthor{\bsnm{Le}, \binits{T.}},
\bauthor{\bsnm{La}, \binits{H.M.}},
\bauthor{\bsnm{Schmid}, \binits{R.}},
\bauthor{\bsnm{Berendsen}, \binits{T.}}:
\bctitle{A multi-functional inspection robot for civil infrastructure evaluation and maintenance}.
In: \bbtitle{2017 {IEEE}/{RSJ} {International} {Conference} on {Intelligent} {Robots} and {Systems} ({IROS})},
pp. \bfpage{2672}--\blpage{2677}
(\byear{2017}).
\doiurl{10.1109/IROS.2017.8206091} .
\bcomment{ISSN: 2153-0866}.
\burl{https://ieeexplore-ieee-org.unr.idm.oclc.org/document/8206091}
Accessed 2024-06-05
\end{bchapter}
\endbibitem

\bibitem[\protect\citeauthoryear{Fragapane et~al.}{}]{fragapane_planning_2021}
\begin{botherref}
\oauthor{\bsnm{Fragapane}, \binits{G.}},
\oauthor{\bsnm{Koster}, \binits{R.}},
\oauthor{\bsnm{Sgarbossa}, \binits{F.}},
\oauthor{\bsnm{Strandhagen}, \binits{J.O.}}:
Planning and control of autonomous mobile robots for intralogistics: Literature review and research agenda
\textbf{294}(2),
405--426
\doiurl{10.1016/j.ejor.2021.01.019} .
Accessed 2023-08-05
\end{botherref}
\endbibitem

\bibitem[\protect\citeauthoryear{Campbell et~al.}{}]{campbell_path_2020}
\begin{botherref}
\oauthor{\bsnm{Campbell}, \binits{S.}},
\oauthor{\bsnm{O'Mahony}, \binits{N.}},
\oauthor{\bsnm{Carvalho}, \binits{A.}},
\oauthor{\bsnm{Krpalkova}, \binits{L.}},
\oauthor{\bsnm{Riordan}, \binits{D.}},
\oauthor{\bsnm{Walsh}, \binits{J.}}:
Path planning techniques for mobile robots a review.
In: 2020 6th International Conference on Mechatronics and Robotics Engineering ({ICMRE}),
pp. 12--16.
\doiurl{10.1109/ICMRE49073.2020.9065187}
\end{botherref}
\endbibitem

\bibitem[\protect\citeauthoryear{Kamon et~al.}{}]{kamon_new_1996}
\begin{botherref}
\oauthor{\bsnm{Kamon}, \binits{I.}},
\oauthor{\bsnm{Rivlin}, \binits{E.}},
\oauthor{\bsnm{Rimon}, \binits{E.}}:
A new range-sensor based globally convergent navigation algorithm for mobile robots.
In: Proceedings of {IEEE} International Conference on Robotics and Automation,
vol. 1,
pp. 429--4351.
\doiurl{10.1109/ROBOT.1996.503814} .
{ISSN}: 1050-4729
\end{botherref}
\endbibitem

\bibitem[\protect\citeauthoryear{Borenstein and Koren}{}]{borenstein_vector_1991}
\begin{botherref}
\oauthor{\bsnm{Borenstein}, \binits{J.}},
\oauthor{\bsnm{Koren}, \binits{Y.}}:
The vector field histogram-fast obstacle avoidance for mobile robots
\textbf{7}(3),
278--288
\doiurl{10.1109/70.88137} .
Conference Name: {IEEE} Transactions on Robotics and Automation
\end{botherref}
\endbibitem

\bibitem[\protect\citeauthoryear{Ulrich and Borenstein}{}]{ulrich_vfhsup_2000}
\begin{botherref}
\oauthor{\bsnm{Ulrich}, \binits{I.}},
\oauthor{\bsnm{Borenstein}, \binits{J.}}:
{VFH}/sup */: local obstacle avoidance with look-ahead verification.
In: Proceedings 2000 {ICRA}. Millennium Conference. {IEEE} International Conference on Robotics and Automation. Symposia Proceedings (Cat. No.00CH37065),
vol. 3,
pp. 2505--25113.
\doiurl{10.1109/ROBOT.2000.846405} .
{ISSN}: 1050-4729
\end{botherref}
\endbibitem

\bibitem[\protect\citeauthoryear{Woods and La}{2019}]{woods_novel_2019}
\begin{barticle}
\bauthor{\bsnm{Woods}, \binits{A.C.}},
\bauthor{\bsnm{La}, \binits{H.M.}}:
\batitle{A {Novel} {Potential} {Field} {Controller} for {Use} on {Aerial} {Robots}}.
\bjtitle{IEEE Transactions on Systems, Man, and Cybernetics: Systems}
\bvolume{49}(\bissue{4}),
\bfpage{665}--\blpage{676}
(\byear{2019})
\doiurl{10.1109/TSMC.2017.2702701} .
Accessed 2024-06-05
\end{barticle}
\endbibitem

\bibitem[\protect\citeauthoryear{Dang et~al.}{2019}]{dang_formation_2019}
\begin{barticle}
\bauthor{\bsnm{Dang}, \binits{A.D.}},
\bauthor{\bsnm{La}, \binits{H.M.}},
\bauthor{\bsnm{Nguyen}, \binits{T.}},
\bauthor{\bsnm{Horn}, \binits{J.}}:
\batitle{Formation control for autonomous robots with collision and obstacle avoidance using a rotational and repulsive force–based approach}.
\bjtitle{International Journal of Advanced Robotic Systems}
\bvolume{16}(\bissue{3}),
\bfpage{1729881419847897}
(\byear{2019})
\doiurl{10.1177/1729881419847897} .
\bcomment{Publisher: SAGE Publications}.
Accessed 2024-06-06
\end{barticle}
\endbibitem

\bibitem[\protect\citeauthoryear{Keith and La}{2024}]{keith_Dynamic_2024}
\begin{bchapter}
\bauthor{\bsnm{Keith}, \binits{R.}},
\bauthor{\bsnm{La}, \binits{H.M.}}:
\bctitle{Dynamic zoning of industrial environments with autonomous mobile robots}.
In: \bbtitle{2024 IEEE Transactions on Automation Science and Engineering}.
\bpublisher{IEEE},
\blocation{Naples, Italy}
(\byear{2024})
\end{bchapter}
\endbibitem

\bibitem[\protect\citeauthoryear{Injarapu and Gawre}{}]{injarapu_survey_2017}
\begin{botherref}
\oauthor{\bsnm{Injarapu}, \binits{A.S.H.H.V.}},
\oauthor{\bsnm{Gawre}, \binits{S.K.}}:
A survey of autonomous mobile robot path planning approaches.
In: 2017 International Conference on Recent Innovations in Signal Processing and Embedded Systems ({RISE}),
pp. 624--628.
\doiurl{10.1109/RISE.2017.8378228}
\end{botherref}
\endbibitem

\bibitem[\protect\citeauthoryear{Zhou et~al.}{}]{zhou_agv_2022}
\begin{botherref}
\oauthor{\bsnm{Zhou}, \binits{W.}},
\oauthor{\bsnm{Qin}, \binits{S.}},
\oauthor{\bsnm{Zhou}, \binits{C.}}:
{AGV} path planning based on improved adaptive genetic algorithm.
In: 2022 International Conference on Artificial Intelligence and Computer Information Technology ({AICIT}),
pp. 1--4.
\doiurl{10.1109/AICIT55386.2022.9930180}
\end{botherref}
\endbibitem

\bibitem[\protect\citeauthoryear{Zadeh}{}]{zadeh_fuzzy_1965}
\begin{botherref}
\oauthor{\bsnm{Zadeh}, \binits{L.A.}}:
Fuzzy sets
\textbf{8}(3),
338--353
\doiurl{10.1016/S0019-9958(65)90241-X} .
Accessed 2023-08-10
\end{botherref}
\endbibitem

\bibitem[\protect\citeauthoryear{Hassanzadeh and Sadigh}{}]{hassanzadeh_path_2009}
\begin{botherref}
\oauthor{\bsnm{Hassanzadeh}, \binits{I.}},
\oauthor{\bsnm{Sadigh}, \binits{S.M.}}:
Path planning for a mobile robot using fuzzy logic controller tuned by {GA}.
In: 2009 6th International Symposium on Mechatronics and Its Applications,
pp. 1--5.
\doiurl{10.1109/ISMA.2009.5164798}
\end{botherref}
\endbibitem

\bibitem[\protect\citeauthoryear{Song et~al.}{}]{song_dynamic_2020}
\begin{botherref}
\oauthor{\bsnm{Song}, \binits{Q.}},
\oauthor{\bsnm{Zhao}, \binits{Q.}},
\oauthor{\bsnm{Wang}, \binits{S.}},
\oauthor{\bsnm{Liu}, \binits{Q.}},
\oauthor{\bsnm{Chen}, \binits{X.}}:
Dynamic path planning for unmanned vehicles based on fuzzy logic and improved ant colony optimization
\textbf{8},
62107--62115
\doiurl{10.1109/ACCESS.2020.2984695} .
Conference Name: {IEEE} Access
\end{botherref}
\endbibitem

\bibitem[\protect\citeauthoryear{Young and La}{2020}]{young_consensus_2020}
\begin{barticle}
\bauthor{\bsnm{Young}, \binits{Z.}},
\bauthor{\bsnm{La}, \binits{H.M.}}:
\batitle{Consensus, cooperative learning, and flocking for multiagent predator avoidance}.
\bjtitle{International Journal of Advanced Robotic Systems}
\bvolume{17}(\bissue{5}),
\bfpage{1729881420960342}
(\byear{2020})
\doiurl{10.1177/1729881420960342} .
\bcomment{Publisher: SAGE Publications}.
Accessed 2024-06-06
\end{barticle}
\endbibitem

\bibitem[\protect\citeauthoryear{Olfati-Saber et~al.}{2007}]{olfati-saber_consensus_2007}
\begin{barticle}
\bauthor{\bsnm{Olfati-Saber}, \binits{R.}},
\bauthor{\bsnm{Fax}, \binits{J.A.}},
\bauthor{\bsnm{Murray}, \binits{R.M.}}:
\batitle{Consensus and {Cooperation} in {Networked} {Multi}-{Agent} {Systems}}.
\bjtitle{Proceedings of the IEEE}
\bvolume{95}(\bissue{1}),
\bfpage{215}--\blpage{233}
(\byear{2007})
\doiurl{10.1109/JPROC.2006.887293} .
Accessed 2024-06-06
\end{barticle}
\endbibitem

\bibitem[\protect\citeauthoryear{Fan et~al.}{}]{fan_research_2021}
\begin{botherref}
\oauthor{\bsnm{Fan}, \binits{M.}},
\oauthor{\bsnm{He}, \binits{J.}},
\oauthor{\bsnm{Ding}, \binits{S.}},
\oauthor{\bsnm{Ding}, \binits{Y.}},
\oauthor{\bsnm{Li}, \binits{M.}},
\oauthor{\bsnm{Jiang}, \binits{L.}}:
Research and implementation of multi-robot path planning based on genetic algorithm.
In: 2021 5th International Conference on Automation, Control and Robots ({ICACR}),
pp. 140--144.
\doiurl{10.1109/ICACR53472.2021.9605194}
\end{botherref}
\endbibitem

\bibitem[\protect\citeauthoryear{Pradhan et~al.}{}]{pradhan_novel_2020}
\begin{botherref}
\oauthor{\bsnm{Pradhan}, \binits{B.}},
\oauthor{\bsnm{Nandi}, \binits{A.}},
\oauthor{\bsnm{Hui}, \binits{N.B.}},
\oauthor{\bsnm{Roy}, \binits{D.S.}},
\oauthor{\bsnm{Rodrigues}, \binits{J.J.P.C.}}:
A novel hybrid neural network-based multirobot path planning with motion coordination
\textbf{69}(2),
1319--1327
\doiurl{10.1109/TVT.2019.2958197} .
Conference Name: {IEEE} Transactions on Vehicular Technology
\end{botherref}
\endbibitem

\bibitem[\protect\citeauthoryear{Li et~al.}{}]{li_graph_2020}
\begin{botherref}
\oauthor{\bsnm{Li}, \binits{Q.}},
\oauthor{\bsnm{Gama}, \binits{F.}},
\oauthor{\bsnm{Ribeiro}, \binits{A.}},
\oauthor{\bsnm{Prorok}, \binits{A.}}:
Graph neural networks for decentralized multi-robot path planning.
In: 2020 {IEEE}/{RSJ} International Conference on Intelligent Robots and Systems ({IROS}),
pp. 11785--11792.
\doiurl{10.1109/IROS45743.2020.9341668} .
{ISSN}: 2153-0866
\end{botherref}
\endbibitem

\bibitem[\protect\citeauthoryear{Chen et~al.}{}]{chen_decentralized_2017}
\begin{botherref}
\oauthor{\bsnm{Chen}, \binits{Y.F.}},
\oauthor{\bsnm{Liu}, \binits{M.}},
\oauthor{\bsnm{Everett}, \binits{M.}},
\oauthor{\bsnm{How}, \binits{J.P.}}:
Decentralized non-communicating multiagent collision avoidance with deep reinforcement learning.
In: 2017 {IEEE} International Conference on Robotics and Automation ({ICRA}),
pp. 285--292.
\doiurl{10.1109/ICRA.2017.7989037}
\end{botherref}
\endbibitem

\bibitem[\protect\citeauthoryear{Ladel}{}]{ladel_common_2022}
\begin{botherref}
\oauthor{\bsnm{Ladel}, \binits{C.}}:
Common Warehouse Management Problems and Their Solutions.
\url{https://www.globaltrademag.com/common-warehouse-management-problems-and-their-solutions/}
Accessed 2023-08-07
\end{botherref}
\endbibitem

\bibitem[\protect\citeauthoryear{Zou et~al.}{}]{zou_evaluating_2018}
\begin{botherref}
\oauthor{\bsnm{Zou}, \binits{B.}},
\oauthor{\bsnm{Xu}, \binits{X.}},
\oauthor{\bsnm{Gong}, \binits{Y.Y.}},
\oauthor{\bsnm{De~Koster}, \binits{R.}}:
Evaluating battery charging and swapping strategies in a robotic mobile fulfillment system
\textbf{267}(2),
733--753
\doiurl{10.1016/j.ejor.2017.12.008} .
Accessed 2023-08-08
\end{botherref}
\endbibitem

\bibitem[\protect\citeauthoryear{Mu et~al.}{}]{mu_battery_2022}
\begin{botherref}
\oauthor{\bsnm{Mu}, \binits{Y.}},
\oauthor{\bsnm{Li}, \binits{Y.}},
\oauthor{\bsnm{Lin}, \binits{K.}},
\oauthor{\bsnm{Deng}, \binits{K.}},
\oauthor{\bsnm{Liu}, \binits{Q.}}:
Battery management for warehouse robots via average-reward reinforcement learning.
In: 2022 {IEEE} International Conference on Robotics and Biomimetics ({ROBIO}),
pp. 253--258.
\doiurl{10.1109/ROBIO55434.2022.10011784}
\end{botherref}
\endbibitem

\bibitem[\protect\citeauthoryear{De~Ryck et~al.}{}]{de_ryck_resource_2020}
\begin{botherref}
\oauthor{\bsnm{De~Ryck}, \binits{M.}},
\oauthor{\bsnm{Versteyhe}, \binits{M.}},
\oauthor{\bsnm{Shariatmadar}, \binits{K.}}:
Resource management in decentralized industrial automated guided vehicle systems
\textbf{54},
204--214
\doiurl{10.1016/j.jmsy.2019.11.003} .
Accessed 2023-08-06
\end{botherref}
\endbibitem

\bibitem[\protect\citeauthoryear{}{}]{noauthor_centralized_nodate}
\begin{botherref}
Centralized Control with {MiR} Fleet {\textbar} Mobile Industrial Robots.
\url{https://www.mobile-industrial-robots.com/solutions/mir-accessories/mir-fleet/}
Accessed 2023-08-05
\end{botherref}
\endbibitem

\bibitem[\protect\citeauthoryear{Chaikovskaia et~al.}{}]{chaikovskaia_sizing_2021}
\begin{botherref}
\oauthor{\bsnm{Chaikovskaia}, \binits{M.}},
\oauthor{\bsnm{Gayon}, \binits{J.-P.}},
\oauthor{\bsnm{Chebab}, \binits{Z.E.}},
\oauthor{\bsnm{Fauroux}, \binits{J.-C.}}:
Sizing of a fleet of cooperative robots for the transport of homogeneous loads.
In: 2021 {IEEE} 17th International Conference on Automation Science and Engineering ({CASE}),
pp. 1654--1659.
\doiurl{10.1109/CASE49439.2021.9551509} .
{ISSN}: 2161-8089
\end{botherref}
\endbibitem

\bibitem[\protect\citeauthoryear{Choobineh et~al.}{}]{choobineh_fleet_2012}
\begin{botherref}
\oauthor{\bsnm{Choobineh}, \binits{F.F.}},
\oauthor{\bsnm{Asef-Vaziri}, \binits{A.}},
\oauthor{\bsnm{Huang}, \binits{X.}}:
Fleet sizing of automated guided vehicles: a linear programming approach based on closed queuing networks
\textbf{50}(12),
3222--3235
\doiurl{10.1080/00207543.2011.562560} .
Publisher: Taylor \& Francis \_eprint: https://doi.org/10.1080/00207543.2011.562560.
Accessed 2023-08-10
\end{botherref}
\endbibitem

\bibitem[\protect\citeauthoryear{Chawla et~al.}{}]{chawla_material_2019}
\begin{botherref}
\oauthor{\bsnm{Chawla}, \binits{V.K.}},
\oauthor{\bsnm{Chanda}, \binits{A.K.}},
\oauthor{\bsnm{Angra}, \binits{S.}}:
Material handling robots fleet size optimization by a heuristic,
177--188
\doiurl{10.5267/j.jpm.2019.4.002} .
Accessed 2023-08-10
\end{botherref}
\endbibitem

\bibitem[\protect\citeauthoryear{Vivaldini et~al.}{}]{vivaldini_integrated_2016}
\begin{botherref}
\oauthor{\bsnm{Vivaldini}, \binits{K.}},
\oauthor{\bsnm{Rocha}, \binits{L.F.}},
\oauthor{\bsnm{Martarelli}, \binits{N.J.}},
\oauthor{\bsnm{Becker}, \binits{M.}},
\oauthor{\bsnm{Moreira}, \binits{A.P.}}:
Integrated tasks assignment and routing for the estimation of the optimal number of {AGVS}
\textbf{82}(1),
719--736
\doiurl{10.1007/s00170-015-7343-4} .
Accessed 2023-08-11
\end{botherref}
\endbibitem

\bibitem[\protect\citeauthoryear{Rjeb et~al.}{}]{rjeb_sizing_2021}
\begin{botherref}
\oauthor{\bsnm{Rjeb}, \binits{A.}},
\oauthor{\bsnm{Gayon}, \binits{J.-P.}},
\oauthor{\bsnm{Norre}, \binits{S.}}:
Sizing of a heterogeneous fleet of robots in a logistics warehouse.
In: 2021 {IEEE} 17th International Conference on Automation Science and Engineering ({CASE}),
pp. 95--100.
\doiurl{10.1109/CASE49439.2021.9551422} .
{ISSN}: 2161-8089
\end{botherref}
\endbibitem

\bibitem[\protect\citeauthoryear{Boccia et~al.}{2024}]{boccia_exact_2024}
\begin{barticle}
\bauthor{\bsnm{Boccia}, \binits{M.}},
\bauthor{\bsnm{Mancuso}, \binits{A.}},
\bauthor{\bsnm{Masone}, \binits{A.}},
\bauthor{\bsnm{Murino}, \binits{T.}},
\bauthor{\bsnm{Sterle}, \binits{C.}}:
\batitle{Exact and heuristic solution approaches for the multi-objective {AGV} scheduling problem with battery constraints}.
\bjtitle{Transportation Research Procedia}
\bvolume{78},
\bfpage{369}--\blpage{376}
(\byear{2024})
\doiurl{10.1016/j.trpro.2024.02.047} .
Accessed 2024-06-07
\end{barticle}
\endbibitem

\bibitem[\protect\citeauthoryear{}{}]{noauthor_fleet_nodate}
\begin{botherref}
Fleet Management System.
\url{https://addverb.com/software/fleet-management-system/}
Accessed 2023-08-07
\end{botherref}
\endbibitem

\bibitem[\protect\citeauthoryear{}{}]{noauthor_what_nodate}
\begin{botherref}
What Is A Warehouse Management System ({WMS})? – Forbes Advisor.
\url{https://www.forbes.com/advisor/business/software/what-is-wms/}
Accessed 2023-08-07
\end{botherref}
\endbibitem

\bibitem[\protect\citeauthoryear{}{}]{noauthor_sap_nodate}
\begin{botherref}
{SAP} Extended Warehouse Management {\textbar} {WMS}.
\url{https://www.sap.com/products/scm/extended-warehouse-management.html}
Accessed 2023-08-08
\end{botherref}
\endbibitem

\bibitem[\protect\citeauthoryear{Anghinolfi et~al.}{}]{anghinolfi_bi-objective_2021}
\begin{botherref}
\oauthor{\bsnm{Anghinolfi}, \binits{D.}},
\oauthor{\bsnm{Paolucci}, \binits{M.}},
\oauthor{\bsnm{Ronco}, \binits{R.}}:
A bi-objective heuristic approach for green identical parallel machine scheduling
\textbf{289}(2),
416--434
\doiurl{10.1016/j.ejor.2020.07.020} .
Accessed 2023-08-12
\end{botherref}
\endbibitem

\bibitem[\protect\citeauthoryear{Yokota}{}]{yokota_min-max-strategy-based_2019}
\begin{botherref}
\oauthor{\bsnm{Yokota}, \binits{T.}}:
Min-max-strategy-based optimum co-operative picking with {AGVs} in warehouse.
In: 2019 58th Annual Conference of the Society of Instrument and Control Engineers of Japan ({SICE}),
pp. 236--242.
\doiurl{10.23919/SICE.2019.8859959}
\end{botherref}
\endbibitem

\bibitem[\protect\citeauthoryear{Yu et~al.}{}]{yu_multi-load_2021}
\begin{botherref}
\oauthor{\bsnm{Yu}, \binits{N.}},
\oauthor{\bsnm{Li}, \binits{T.}},
\oauthor{\bsnm{Wang}, \binits{B.}}:
Multi-load {AGVs} scheduling algorithm in automated sorting warehouse.
In: 2021 14th International Symposium on Computational Intelligence and Design ({ISCID}),
pp. 126--129.
\doiurl{10.1109/ISCID52796.2021.00037} .
{ISSN}: 2473-3547
\end{botherref}
\endbibitem

\bibitem[\protect\citeauthoryear{Tang et~al.}{}]{tang_novel_2021}
\begin{botherref}
\oauthor{\bsnm{Tang}, \binits{H.}},
\oauthor{\bsnm{Wang}, \binits{A.}},
\oauthor{\bsnm{Xue}, \binits{F.}},
\oauthor{\bsnm{Yang}, \binits{J.}},
\oauthor{\bsnm{Cao}, \binits{Y.}}:
A novel hierarchical soft actor-critic algorithm for multi-logistics robots task allocation
\textbf{9},
42568--42582
\doiurl{10.1109/ACCESS.2021.3062457} .
Conference Name: {IEEE} Access
\end{botherref}
\endbibitem

\bibitem[\protect\citeauthoryear{Li and Wu}{}]{li_research_2022}
\begin{botherref}
\oauthor{\bsnm{Li}, \binits{C.}},
\oauthor{\bsnm{Wu}, \binits{W.}}:
Research on cooperative scheduling of {AGV} transportation and charging in intelligent warehouse system based on dynamic task chain.
In: 2022 {IEEE} International Conference on Networking, Sensing and Control ({ICNSC}),
pp. 1--6.
\doiurl{10.1109/ICNSC55942.2022.10004100}
\end{botherref}
\endbibitem

\bibitem[\protect\citeauthoryear{Li and Huang}{2024}]{li_efficient_2024}
\begin{botherref}
\oauthor{\bsnm{Li}, \binits{Y.}},
\oauthor{\bsnm{Huang}, \binits{H.}}:
Efficient {Task} {Planning} for {Heterogeneous} {AGVs} in {Warehouses}.
IEEE Transactions on Intelligent Transportation Systems,
1--15
(2024)
\doiurl{10.1109/TITS.2024.3356514} .
Conference Name: IEEE Transactions on Intelligent Transportation Systems.
Accessed 2024-06-07
\end{botherref}
\endbibitem

\bibitem[\protect\citeauthoryear{Basile et~al.}{}]{basile_auction-based_2017}
\begin{botherref}
\oauthor{\bsnm{Basile}, \binits{F.}},
\oauthor{\bsnm{Chiacchio}, \binits{P.}},
\oauthor{\bsnm{Di~Marino}, \binits{E.}}:
An auction-based approach for the coordination of vehicles in automated warehouse systems.
In: 2017 {IEEE} International Conference on Service Operations and Logistics, and Informatics ({SOLI}),
pp. 121--126.
\doiurl{10.1109/SOLI.2017.8120981}
\end{botherref}
\endbibitem

\bibitem[\protect\citeauthoryear{Warita and Fujita}{2024}]{warita_online_2024}
\begin{barticle}
\bauthor{\bsnm{Warita}, \binits{S.}},
\bauthor{\bsnm{Fujita}, \binits{K.}}:
\batitle{Online {Planning} for {Autonomous} {Mobile} {Robots} with {Different} {Objectives} in {Warehouse} {Commissioning} {Task}}.
\bjtitle{Information}
\bvolume{15}(\bissue{3}),
\bfpage{130}
(\byear{2024})
\doiurl{10.3390/info15030130} .
\bcomment{Number: 3 Publisher: Multidisciplinary Digital Publishing Institute}.
Accessed 2024-06-07
\end{barticle}
\endbibitem

\bibitem[\protect\citeauthoryear{de~Koster et~al.}{}]{de_koster_design_2007}
\begin{botherref}
\oauthor{\bsnm{Koster}, \binits{R.}},
\oauthor{\bsnm{Le-Duc}, \binits{T.}},
\oauthor{\bsnm{Roodbergen}, \binits{K.J.}}:
Design and control of warehouse order picking: A literature review
\textbf{182}(2),
481--501
\doiurl{10.1016/j.ejor.2006.07.009} .
Accessed 2023-08-29
\end{botherref}
\endbibitem

\bibitem[\protect\citeauthoryear{Boysen et~al.}{}]{boysen_parts--picker_2017}
\begin{botherref}
\oauthor{\bsnm{Boysen}, \binits{N.}},
\oauthor{\bsnm{Briskorn}, \binits{D.}},
\oauthor{\bsnm{Emde}, \binits{S.}}:
Parts-to-picker based order processing in a rack-moving mobile robots environment
\textbf{262}(2),
550--562
\doiurl{10.1016/j.ejor.2017.03.053} .
Accessed 2023-08-29
\end{botherref}
\endbibitem

\bibitem[\protect\citeauthoryear{Srinivas and Yu}{}]{srinivas_collaborative_2022}
\begin{botherref}
\oauthor{\bsnm{Srinivas}, \binits{S.}},
\oauthor{\bsnm{Yu}, \binits{S.}}:
Collaborative order picking with multiple pickers and robots: Integrated approach for order batching, sequencing and picker-robot routing
\textbf{254},
108634
\doiurl{10.1016/j.ijpe.2022.108634} .
Accessed 2023-08-30
\end{botherref}
\endbibitem

\bibitem[\protect\citeauthoryear{Žulj et~al.}{}]{zulj_order_2022}
\begin{botherref}
\oauthor{\bsnm{Žulj}, \binits{I.}},
\oauthor{\bsnm{Salewski}, \binits{H.}},
\oauthor{\bsnm{Goeke}, \binits{D.}},
\oauthor{\bsnm{Schneider}, \binits{M.}}:
Order batching and batch sequencing in an {AMR}-assisted picker-to-parts system
\textbf{298}(1),
182--201
\doiurl{10.1016/j.ejor.2021.05.033} .
Accessed 2023-08-30
\end{botherref}
\endbibitem

\bibitem[\protect\citeauthoryear{Zhao et~al.}{2024}]{zhao_order_2024}
\begin{barticle}
\bauthor{\bsnm{Zhao}, \binits{Z.}},
\bauthor{\bsnm{Cheng}, \binits{J.}},
\bauthor{\bsnm{Liang}, \binits{J.}},
\bauthor{\bsnm{Liu}, \binits{S.}},
\bauthor{\bsnm{Zhou}, \binits{M.}},
\bauthor{\bsnm{Al-Turki}, \binits{Y.}}:
\batitle{Order {Picking} {Optimization} in {Smart} {Warehouses} {With} {Human}–{Robot} {Collaboration}}.
\bjtitle{IEEE Internet of Things Journal}
\bvolume{11}(\bissue{9}),
\bfpage{16314}--\blpage{16324}
(\byear{2024})
\doiurl{10.1109/JIOT.2024.3352658} .
\bcomment{Conference Name: IEEE Internet of Things Journal}.
Accessed 2024-06-10
\end{barticle}
\endbibitem

\bibitem[\protect\citeauthoryear{Wan and Liu}{2022}]{wan_integrating_2022}
\begin{botherref}
\oauthor{\bsnm{Wan}, \binits{Y.}},
\oauthor{\bsnm{Liu}, \binits{Y.}}:
Integrating {Optimized} {Fishbone} {Warehouse} {Layout}, {Storage} {Location} {Assignment} and {Picker} {Routing}. {\textbar} {IAENG} {International} {Journal} of {Computer} {Science} {\textbar} {EBSCOhost}.
ISSN: 1819-656X Issue: 3 Pages: 957 Volume: 49
(2022).
\url{https://openurl.ebsco.com/contentitem/gcd:158904002?sid=ebsco:plink:crawler&id=ebsco:gcd:158904002}
Accessed 2024-06-11
\end{botherref}
\endbibitem

\bibitem[\protect\citeauthoryear{Masae et~al.}{}]{masae_order_2020}
\begin{botherref}
\oauthor{\bsnm{Masae}, \binits{M.}},
\oauthor{\bsnm{Glock}, \binits{C.H.}},
\oauthor{\bsnm{Grosse}, \binits{E.H.}}:
Order picker routing in warehouses: A systematic literature review
\textbf{224},
107564
\doiurl{10.1016/j.ijpe.2019.107564} .
Accessed 2023-09-01
\end{botherref}
\endbibitem

\bibitem[\protect\citeauthoryear{Scholz and Wäscher}{}]{scholz_order_2017}
\begin{botherref}
\oauthor{\bsnm{Scholz}, \binits{A.}},
\oauthor{\bsnm{Wäscher}, \binits{G.}}:
Order batching and picker routing in manual order picking systems: the benefits of integrated routing
\textbf{25}(2),
491--520
\doiurl{10.1007/s10100-017-0467-x} .
Accessed 2023-09-02
\end{botherref}
\endbibitem

\bibitem[\protect\citeauthoryear{Cano et~al.}{}]{cano_solving_2023}
\begin{botherref}
\oauthor{\bsnm{Cano}, \binits{J.A.}},
\oauthor{\bsnm{Cortés}, \binits{P.}},
\oauthor{\bsnm{Muñuzuri}, \binits{J.}},
\oauthor{\bsnm{Correa-Espinal}, \binits{A.}}:
Solving the picker routing problem in multi-block high-level storage systems using metaheuristics
\textbf{35}(2),
376--415
\doiurl{10.1007/s10696-022-09445-y} .
Accessed 2023-09-01
\end{botherref}
\endbibitem

\bibitem[\protect\citeauthoryear{Manzini}{}]{manzini_warehousing_2012}
\begin{botherref}
\oauthor{\bsnm{Manzini}, \binits{R.}}:
Warehousing in the Global Supply Chain: Advanced Models, Tools and Applications for Storage Systems.
Springer.
Google-Books-{ID}: {EZp}1iiVEo9MC
\end{botherref}
\endbibitem

\bibitem[\protect\citeauthoryear{Roodbergen et~al.}{}]{roodbergen_designing_2008}
\begin{botherref}
\oauthor{\bsnm{Roodbergen}, \binits{K.J.}},
\oauthor{\bsnm{Sharp}, \binits{G.P.}},
\oauthor{\bsnm{Vis}, \binits{I.F.A.}}:
Designing the layout structure of manual order picking areas in warehouses
\textbf{40}(11),
1032--1045
\doiurl{10.1080/07408170802167639} .
Publisher: Taylor \& Francis \_eprint: https://doi.org/10.1080/07408170802167639.
Accessed 2023-09-02
\end{botherref}
\endbibitem

\bibitem[\protect\citeauthoryear{Diefenbach and Glock}{}]{diefenbach_ergonomic_2019}
\begin{botherref}
\oauthor{\bsnm{Diefenbach}, \binits{H.}},
\oauthor{\bsnm{Glock}, \binits{C.H.}}:
Ergonomic and economic optimization of layout and item assignment of a u-shaped order picking zone
\textbf{138},
106094
\doiurl{10.1016/j.cie.2019.106094} .
Accessed 2023-09-01
\end{botherref}
\endbibitem

\bibitem[\protect\citeauthoryear{Hamzeei et~al.}{}]{hamzeei_exact_2013}
\begin{botherref}
\oauthor{\bsnm{Hamzeei}, \binits{M.}},
\oauthor{\bsnm{Farahani}, \binits{R.Z.}},
\oauthor{\bsnm{Rashidi-Bejgan}, \binits{H.}}:
An exact and a simulated annealing algorithm for simultaneously determining flow path and the location of p/d stations in bidirectional path
\textbf{32}(4),
648--654
\doiurl{10.1016/j.jmsy.2013.07.002} .
Accessed 2023-08-11
\end{botherref}
\endbibitem

\bibitem[\protect\citeauthoryear{Li and Li}{}]{li_bi-objective_2023}
\begin{botherref}
\oauthor{\bsnm{Li}, \binits{Y.}},
\oauthor{\bsnm{Li}, \binits{Z.}}:
Bi-objective optimization for multi-row facility layout problem integrating automated guided vehicle path
\textbf{11},
55954--55964
\doiurl{10.1109/ACCESS.2023.3281554} .
Conference Name: {IEEE} Access
\end{botherref}
\endbibitem

\bibitem[\protect\citeauthoryear{Keller and Buscher}{}]{keller_single_2015}
\begin{botherref}
\oauthor{\bsnm{Keller}, \binits{B.}},
\oauthor{\bsnm{Buscher}, \binits{U.}}:
Single row layout models
\textbf{245}(3),
629--644
\doiurl{10.1016/j.ejor.2015.03.016} .
Accessed 2023-08-11
\end{botherref}
\endbibitem

\bibitem[\protect\citeauthoryear{Zhang et~al.}{2023}]{zhang_multi-robot_2023}
\begin{bbook}
\bauthor{\bsnm{Zhang}, \binits{Y.}},
\bauthor{\bsnm{Fontaine}, \binits{M.}},
\bauthor{\bsnm{Bhatt}, \binits{V.}},
\bauthor{\bsnm{Nikolaidis}, \binits{S.}},
\bauthor{\bsnm{Li}, \binits{J.}}:
\bbtitle{Multi-{Robot} {Coordination} and {Layout} {Design} for {Automated} {Warehousing}},
(\byear{2023})
\end{bbook}
\endbibitem

\bibitem[\protect\citeauthoryear{Bao et~al.}{}]{bao_storage_2019}
\begin{botherref}
\oauthor{\bsnm{Bao}, \binits{L.G.}},
\oauthor{\bsnm{Dang}, \binits{T.G.}},
\oauthor{\bsnm{Anh}, \binits{N.D.}}:
Storage assignment policy and route planning of {AGVS} in warehouse optimization.
In: 2019 International Conference on System Science and Engineering ({ICSSE}),
pp. 599--604.
\doiurl{10.1109/ICSSE.2019.8823418} .
{ISSN}: 2325-0925
\end{botherref}
\endbibitem

\bibitem[\protect\citeauthoryear{Wessling}{}]{wessling_chery_2023}
\begin{botherref}
\oauthor{\bsnm{Wessling}, \binits{B.}}:
Chery Automotive Deploys 100+ {ForwardX} {AMRs} in Super One Factory.
\url{https://mobilerobotguide.com/2023/07/14/chery-automotive-deploys-100-forwardx-amrs-in-super-one-factory/}
Accessed 2023-08-13
\end{botherref}
\endbibitem

\bibitem[\protect\citeauthoryear{Gehlhoff and Fay}{}]{gehlhoff_agent-based_2020}
\begin{botherref}
\oauthor{\bsnm{Gehlhoff}, \binits{F.}},
\oauthor{\bsnm{Fay}, \binits{A.}}:
Agent-based decentralised architecture for multi-stage and integrated scheduling.
In: 2020 25th {IEEE} International Conference on Emerging Technologies and Factory Automation ({ETFA}),
vol. 1,
pp. 1443--1446.
\doiurl{10.1109/ETFA46521.2020.9212059} .
{ISSN}: 1946-0759
\end{botherref}
\endbibitem

\bibitem[\protect\citeauthoryear{Chen and Tiong}{}]{chen_using_2018}
\begin{botherref}
\oauthor{\bsnm{Chen}, \binits{C.}},
\oauthor{\bsnm{Tiong}, \binits{R.}}:
Using queuing theory and simulated annealing to design the facility layout in an {AGV}-based modular manufacturing system
\textbf{57},
1--18
\doiurl{10.1080/00207543.2018.1533654}
\end{botherref}
\endbibitem

\bibitem[\protect\citeauthoryear{Ho and Liao}{}]{ho_zone_2009}
\begin{botherref}
\oauthor{\bsnm{Ho}, \binits{Y.-C.}},
\oauthor{\bsnm{Liao}, \binits{T.-W.}}:
Zone design and control for vehicle collision prevention and load balancing in a zone control {AGV} system
\textbf{56}(1),
417--432
\doiurl{10.1016/j.cie.2008.07.007} .
Accessed 2023-08-12
\end{botherref}
\endbibitem

\bibitem[\protect\citeauthoryear{Saylam et~al.}{}]{saylam_minmax_2023}
\begin{botherref}
\oauthor{\bsnm{Saylam}, \binits{S.}},
\oauthor{\bsnm{Çelik}, \binits{M.}},
\oauthor{\bsnm{Süral}, \binits{H.}}:
The min–max order picking problem in synchronised dynamic zone-picking systems
\textbf{61}(7),
2086--2104
\doiurl{10.1080/00207543.2022.2058433} .
Publisher: Taylor \& Francis \_eprint: https://doi.org/10.1080/00207543.2022.2058433.
Accessed 2023-08-12
\end{botherref}
\endbibitem

\bibitem[\protect\citeauthoryear{Li et~al.}{}]{li_multi-agvs_2019}
\begin{botherref}
\oauthor{\bsnm{Li}, \binits{X.}},
\oauthor{\bsnm{Zhang}, \binits{C.}},
\oauthor{\bsnm{Yang}, \binits{W.}},
\oauthor{\bsnm{Qi}, \binits{M.}}:
Multi-{AGVs} conflict-free routing and dynamic dispatching strategies for automated warehouses.
In: Kim, K.J., Kim, H. (eds.)
Mobile and Wireless Technology 2018.
Lecture Notes in Electrical Engineering,
pp. 277--286.
Springer.
\doiurl{10.1007/978-981-13-1059-1_26}
\end{botherref}
\endbibitem

\bibitem[\protect\citeauthoryear{Yuan and Gong}{}]{yuan_bot--time_2017}
\begin{botherref}
\oauthor{\bsnm{Yuan}, \binits{Z.}},
\oauthor{\bsnm{Gong}, \binits{Y.Y.}}:
Bot-in-time delivery for robotic mobile fulfillment systems
\textbf{64}(1),
83--93
\doiurl{10.1109/TEM.2016.2634540} .
Conference Name: {IEEE} Transactions on Engineering Management
\end{botherref}
\endbibitem

\bibitem[\protect\citeauthoryear{Tang et~al.}{}]{tang_research_2021}
\begin{botherref}
\oauthor{\bsnm{Tang}, \binits{H.}},
\oauthor{\bsnm{Cheng}, \binits{X.}},
\oauthor{\bsnm{Jiang}, \binits{W.}},
\oauthor{\bsnm{Chen}, \binits{S.}}:
Research on equipment configuration optimization of {AGV} unmanned warehouse
\textbf{9},
47946--47959
\doiurl{10.1109/ACCESS.2021.3066622} .
Conference Name: {IEEE} Access
\end{botherref}
\endbibitem

\bibitem[\protect\citeauthoryear{Correia et~al.}{}]{correia_implementing_2020}
\begin{botherref}
\oauthor{\bsnm{Correia}, \binits{N.}},
\oauthor{\bsnm{Teixeira}, \binits{L.}},
\oauthor{\bsnm{Ramos}, \binits{A.L.}}:
Implementing an {AGV} system to transport finished goods to the warehouse
\textbf{5}(2),
241--247
\doiurl{10.25046/aj050231} .
Accessed 2023-08-13
\end{botherref}
\endbibitem

\bibitem[\protect\citeauthoryear{Ilić}{}]{ilic_analysis_1994}
\begin{botherref}
\oauthor{\bsnm{Ilić}, \binits{O.R.}}:
Analysis of the number of automated guided vehicles required in flexible manufacturing systems
\textbf{9}(6),
382--389
\doiurl{10.1007/BF01748483} .
Accessed 2023-08-13
\end{botherref}
\endbibitem

\bibitem[\protect\citeauthoryear{Jaghbeer et~al.}{}]{jaghbeer_automated_2020}
\begin{botherref}
\oauthor{\bsnm{Jaghbeer}, \binits{Y.}},
\oauthor{\bsnm{Hanson}, \binits{R.}},
\oauthor{\bsnm{Johansson}, \binits{M.I.}}:
Automated order picking systems and the links between design and performance: a systematic literature review
\textbf{58}(15),
4489--4505
\doiurl{10.1080/00207543.2020.1788734} .
Publisher: Taylor \& Francis \_eprint: https://doi.org/10.1080/00207543.2020.1788734.
Accessed 2023-08-31
\end{botherref}
\endbibitem

\end{thebibliography}

\end{document}